\pdfoutput=1

\documentclass[11pt]{article}
\usepackage{authblk}
\usepackage{EMNLP2023}

\usepackage{times}
\usepackage{latexsym}

\usepackage[T1]{fontenc}

\usepackage[utf8]{inputenc}

\usepackage{microtype}

\usepackage{inconsolata}
\usepackage{amsmath}

\usepackage{listings}
\usepackage{xcolor}

\usepackage{graphicx}
\usepackage{amsfonts}
\usepackage{svg}
\usepackage{worldflags}
\usepackage{multirow}
\usepackage{comment}
\usepackage{cleveref}
\usepackage{pifont}
\usepackage{array}
\newcommand{\checkMark}{\ding{51}}%
\newcommand{\xMark}{\ding{55}}%

\newcommand{\capivaraicon}[1]{\includegraphics[width=#1]{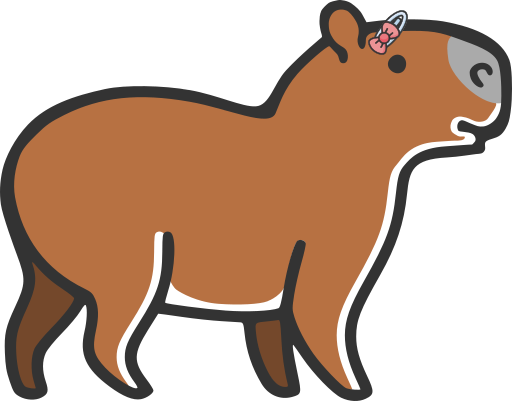}}

\definecolor{codeblue}{rgb}{0.18, 0.35, 0.47}
\definecolor{codegray}{rgb}{0.5,0.5,0.5}
\definecolor{codepurple}{rgb}{0.58,0,0.82}
\definecolor{backcolour}{rgb}{0.95,0.95,0.92}

\lstdefinestyle{mystyle}{
    commentstyle=\color{codeblue},
    keywordstyle=\color{magenta},
    numberstyle=\tiny\color{codegray},
    stringstyle=\color{codepurple},
    basicstyle=\ttfamily\footnotesize,
    breakatwhitespace=false,         
    breaklines=true,                 
    captionpos=b,                    
    keepspaces=true,                 
    showspaces=false,                
    showstringspaces=false,
    showtabs=false,                  
    tabsize=1,
    frame=single, 
    float=!htb
}

\makeatletter
\renewcommand\outauthor{
    \begin{tabular}[t]{>{\centering}p{13cm}}
    \bf\@author
    \end{tabular}}
\makeatother

\title{\capivaraicon{2em} CAPIVARA: Cost-Efficient Approach for Improving Multilingual CLIP Performance on Low-Resource Languages}

\author{Gabriel Oliveira dos Santos$^{1}$\thanks{~Equal contribution. Corresponding authors: G.O.S. (gabriel.santos@ic.unicamp.br), D.A.B.M. (diego.moreira@ ic.unicamp.br) and S.A. (avilas@unicamp.br).}\hspace{0.15cm}, Diego A. B. Moreira$^{1*}$, Alef Iury Ferreira$^{2}$, Jhessica Silva$^{1}$, Luiz Pereira$^{1}$, Pedro Bueno$^{1}$, Thiago Sousa$^{2}$, Helena Maia$^{1}$, Nádia da Silva$^{2}$, Esther Colombini$^{1}$, Helio Pedrini$^{1}$, Sandra Avila$^{1}$\vspace{-0.25cm}\\ 
\hspace{-0.925cm}~$^{1}$Instituto de Computação, Universidade Estadual de Campinas (UNICAMP), Brasil \\
\hspace{-0.5cm}~$^{2}$Instituto de Informática, Universidade Federal de Goiás (UFG), Brasil\\}

\begin{document}
\tolerance=999
\sloppy

\maketitle

\begin{abstract}
This work introduces CAPIVARA, a cost-efficient framework designed to enhance the performance of multilingual CLIP models in low-resource languages. While CLIP has excelled in zero-shot vision-language tasks, the resource-intensive nature of model training remains challenging. Many datasets lack linguistic diversity, featuring solely English descriptions for images. CAPIVARA addresses this by augmenting text data using image captioning and machine translation to generate multiple synthetic captions in low-resource languages. We optimize the training pipeline with LiT, LoRA, and gradient checkpointing to alleviate the computational cost. Through extensive experiments, CAPIVARA emerges as state of the art in zero-shot tasks involving images and Portuguese texts. We show the potential for significant improvements in other low-resource languages, achieved by fine-tuning the pre-trained multilingual CLIP using CAPIVARA on a single GPU for 2 hours. Our model and code is available at \url{https://github.com/hiaac-nlp/CAPIVARA}.
\end{abstract}

\section{Introduction}

The challenge of learning a joint multimodal representation for vision and language has developed various pre-trained models in recent years~\cite{wang2021efficientclip, gao2021clip, yang2022vision, geng2022hiclip, blip2_li_2023}. Remarkably, CLIP~\cite{radford2021learning} has gained attention for achieving state of the art on zero-shot vision-language tasks through contrastive learning to align images and text within a multimodal embedding.

\begin{figure}[t]
  \centering
  \includegraphics[width=\columnwidth, page=1]{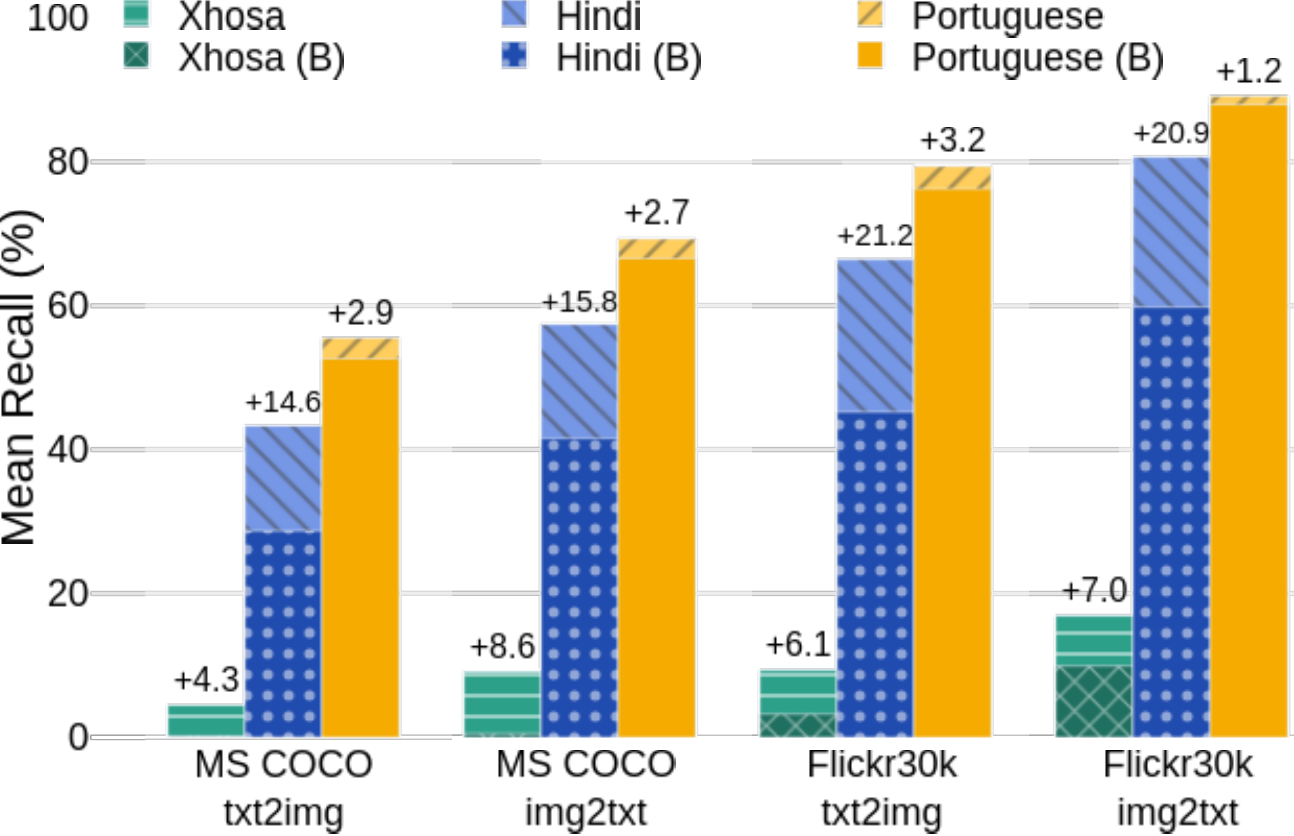}  
  \caption{Improving multilingual CLIP Performance in Low-Resource Languages: Xhosa, Hindi, and Portuguese. This figure illustrates CAPIVARA's effectiveness in enhancing the performance of pre-trained multilingual CLIP models, the \textsc{Open-CLIP} baseline (B), for low-resource languages. The percentage point increase in mean recall for text-to-image (txt2img) and image-to-text (img2txt) retrieval with low-resource languages on Flickr30k and MS COCO datasets is highlighted above the respective bars. CAPIVARA significantly improves the model's baseline performance with only 2 hours of training and 8.5 GB of GPU memory.}
  \label{fig:low-resource-performance}
\end{figure}

Training models such as CLIP requires massive data and computational resources despite their good generalization capacity. These models are typically trained with datasets containing hundreds of millions of image-text pairs, often collected from the web. However, many datasets only provide images paired with English descriptions; as a result, the research community focuses excessively on English texts, whereas other languages are neglected, reinforcing cultural, regional, and linguistic biases~\cite{bender2021dangers}. 
While recent advancements include approaches for languages beyond English~\cite{italianCLIP_bianchi_2021, cn_clip_yang_2022, kelip_Ko_Gu_2022} and multilingual methods~\cite{mCLIP_Carlsson_2022, AltCLIP_chen_2022}, they primarily focus on high-resource languages. There is a scarcity of approaches considering low-resource languages, and even models including them show performance disparities in tasks involving these languages compared to tasks with English~texts.

We propose a \textbf{\underline{c}}ost-efficient \textbf{\underline{ap}}proach for \textbf{\underline{i}}mpro\textbf{\underline{v}}ing multilingual CLIP perform\textbf{\underline{a}}nce in low-\textbf{\underline{r}}esource l\textbf{\underline{a}}nguages (\text{CAPIVARA}), addressing the performance gap with English and reducing computational requirements. Our approach relies on the assumption that datasets may contain images annotated with noisy descriptions. In this way, our framework utilizes BLIP2~\cite{blip2_li_2023} to generate multiple synthetic captions for each image, addressing noisy annotations and limited language diversity challenges. Using the re-annotated dataset, we translate both the original and generated captions into the target language and conduct fine-tuning on the multilingual model. To mitigate the computational cost associated with CLIP model training, we propose to optimize the training pipeline with LiT strategy~\cite{lit_Zhai_2022}, wherein the image encoder remains frozen during training, gradient checkpointing~\cite{chen2016training} and LoRA~\cite{hu2021lora}. Figure~\ref{fig:low-resource-performance} demonstrates that substantial improvements in low-resource language can be achieved by fine-tuning the pre-trained multilingual CLIP with CAPIVARA. 

Our main contributions are as follows:
\begin{itemize}
    \item We introduce CAPIVARA, a low-cost data-centric framework that leverages image captioning models to enhance the annotation of existing datasets to improve the performance of pre-trained multilingual CLIP in low-resource languages. We report the carbon footprint of our method.     
    \item To the best of our knowledge, we are the first to employ LoRA for langua\-ge adaptation in CLIP models, considerably reducing the number of trainable parameters.
    \item We show that augmenting text data, by generating multiple image-conditioned captions with image captioning models, can boost CLIP performance in low-resource language.
    \item We achieve state of the art in many zero-shot tasks involving images and Portuguese texts. This work aims to push forward the multimodal learning literature in the Portuguese-speaking community\footnote{Portuguese, despite being ranked fifth among world languages in the number of native speakers, is a low-resource language from a machine-learning perspective.}.
    \item We make available the re-annotated CC3M with descriptions in Portuguese and English for seamless utilization by other researchers as a data augmentation resource. We also provide the annotations translated to Portuguese for Flickr30k, MS COCO, CC3M, \text{ImageNet-1k}, and ELEVATER datasets.
\end{itemize}

\section{Related Work}

\paragraph{CLIP.}\hspace{-0.15cm}The multimodal vision and language model known as CLIP (Contrastive Language-Image Pre-training) \cite{radford2021learning} rapidly gained attention for its simplicity, scalability, and impressive results. It is pre-trained on 400~million image-text pairs to learn a contrastive representation of images and texts in a multimodal space. 

OpenCLIP~\cite{openclip_2021} is an open-source initiative that provides CLIP models trained on large datasets. It offers well-trained and robust models for pre-training purposes. Based on the original CLIP architecture, OpenCLIP maintains similar accuracy when trained on the same dataset. However, it extends its training to datasets like LAION-400M, LAION-2B, and \text{DataComp-1B}. Unlike the original CLIP, OpenCLIP introduces various image and text encoder configurations, including the \textsc{OpenCLIP ViT-B/32 XLM-Roberta Base} used in this work.
   
\vspace{-0.1cm}\paragraph{Non-English CLIPs.}\hspace{-0.15cm}\citet{italianCLIP_bianchi_2021} introduce the first non-English CLIP-based models. The Italian CLIP model, unlike the original CLIP model, is trained using networks previously pre-trained in text and image tasks. It employs 1.4~million samples from translated datasets. 

The Chinese CLIP \cite{cn_clip_yang_2022} explores different training approaches. The most effective architecture combines a pre-trained model with the LiT (Locked-image text Tuning) strategy~\cite{lit_Zhai_2022}, freezing the text encoder until stability and extensive parameter training. Training data comprises 200~million image-text pairs. 

The Korean CLIP (KELIP) model~\cite{kelip_Ko_Gu_2022} focuses on training from scratch using substantial data and language-specific techniques. It involves self-supervised pre-training of the image encoder and alignment with the English CLIP version. The training dataset comprises 1.1 billion examples, including 708~million Korean samples. 

\vspace{-0.1cm}\paragraph{Multilingual CLIPs.}\hspace{-0.15cm}M-CLIP (Multilingual CLIP) \cite{mCLIP_Carlsson_2022} builds on the pre-trained CLIP model, using its text encoder while discarding the visual encoder. It employs a teacher-learning technique to transfer knowledge from a pre-trained teacher network to new language models. M-CLIP is applied to 68 languages, translated versions of datasets by the MarianMT model~\cite{mariannmt}.

AltCLIP (Altering the Language Encoder in CLIP)~\cite{AltCLIP_chen_2022} introduces a bilingual model for Chinese and a multilingual one for 11~languages. Like M-CLIP, the teacher-learning technique uses only the textual model across various languages. However, AltCLIP differs by incorporating English text distillation, human-curated translations, and a final fine-tuning phase. It also uses the LiT strategy to freeze the image encoder. 

\vspace{-0.1cm}\paragraph{Data-Centric Approaches.}\hspace{-0.15cm}Multimodal learning has been mainly explored through algorithmic designs, often treating datasets as monolithic. \citet{santurkar2023is} reveal that CLIP's performance depends on three pre-training datasets properties: dataset size, caption descriptiveness, and caption variability for each image. They employ BLIP (Bootstrapping Language-Image Pre-training)~\cite{blip_li_2022} to generate new captions to address limited text diversity, improving CLIP performan\-ce. Similarly, \citet{laclip_fan2023improving} propose \text{LaCLIP} (Language augmented CLIP) that uses LLM (Large Language Model) to rewrite captions to increase the text diversity within text-image pairs in the pre-training dataset. However, the decoupled text-generation process might limit effectiveness in datasets with non-descriptive captions~\cite{nguyen2023improving}. 

Our work is related to \citet{laclip_fan2023improving} and \citet{nguyen2023improving}. However, their studies focus on English captions during training and require extensive computational resources. In contrast, our research addresses a constrained scenario with limited computational power --- a single GPU --- and a lack of annotated datasets in the target language. We leverage multilingual \text{OpenCLIP} and English-annotated open datasets to enhance model performance in Portuguese. Our method, centered on Portuguese-translated captions, can be extended to other languages, making it well-suited for low-resource language challenges.

\section{Method}
\label{sec:method}

\begin{figure*}[t]
  \centering
  \includegraphics[width=\textwidth, page=1]{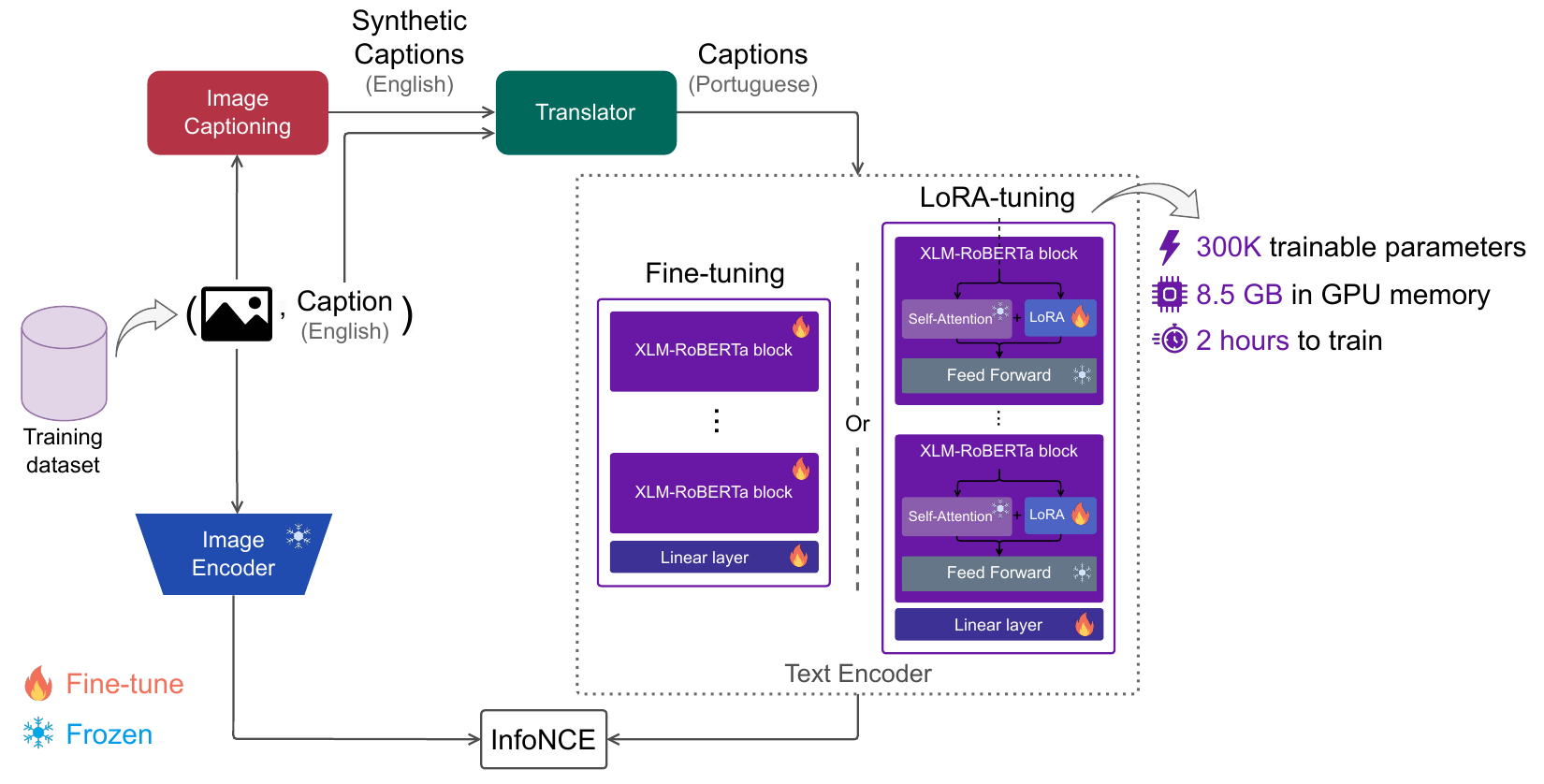}
  \caption{CAPIVARA overview. In our framework, the training dataset comprehends images annotated with English captions. To enhance the annotations, we use an image captioning model to generate synthetic captions for the images. Then, both original and synthetic captions are translated from English to the target language, in our case, Portuguese. We freeze the image encoder and fine-tune the text encoder using the translated captions to align the visual representation by optimizing the InfoNCE loss. While it is possible to fine-tune the entire text encoder, such an approach is resource-intensive. Thus, we propose an optimization method based on LoRA-tuning that can significantly reduce the associated computational cost and speed up the training time.}
  \label{fig:capivara_overview}
\end{figure*}

This section details our approach, including generating captions, translating them into Portuguese, and integrating these new captions into the training pipeline. It also describes optimization through LoRA and gradient checkpointing, effectively reducing the computational resources for CLIP model training. Figure~\ref{fig:capivara_overview} illustrates the main components of CAPIVARA.

\subsection{Model Architecture}
\label{sec:model_architecture}
We use the pre-trained multilingual model \textsc{OpenCLIP ViT-B/32 XLM-Roberta Base}\footnote{\scriptsize{\url{https://huggingface.co/laion/CLIP-ViT-B-32-xlm-roberta-base-laion5B-s13B-b90k}}} (\textsc{OpenCLIP} for short). This model utilizes XLM-RoBERTa Base~\cite{conneau2019unsupervised} and ViT Base~\cite{dosovitskiy2020image} with 32$\times$32 resolution as text and image encoder, respectively. The model was pre-trained on LAION-5B~\cite{schuhmann2022laion} for 12.8B steps and a batch size of~90k. \textcolor{black}{We employ base versions of the encoders, as larger models would demand significantly greater computational resources for both training and inference. This consideration is crucial when addressing the low-resource language community.}

\subsection{Datasets}
\label{sec:datasets}
We use CC3M~\cite{sharma2018conceptual} and modifications over it to fine-tune the \textsc{OpenCLIP} model to improve its performance in Portuguese. For zero-shot text-to-image and image-to-text retrieval tasks, we use PraCegoVer~\cite{pracegover2022}, which is composed of images annotated originally with Portuguese texts, and our Portuguese-translated versions of MS COCO~\cite{MSCoco2014} and Flickr30k~\cite{Flickr30K2017}. We also translate the labels from ImageNet~\cite{deng2009imagenet} and the  ELEVATER benchmark datasets~\cite{li2022elevater} for image classification.

\subsection{Dataset Filtering}
\label{sec:dataset_filtering}
Similar to~\citet{schuhmann2022laion, gadre2023datacomp}, we apply CLIP score filtering. Thus, we discard examples where the cosine similarity, computed by \textsc{OpenCLIP ViT-B/32 XLM-Roberta Base}, between the image and text embeddings is lower than 0.20. We employ this method to CC3M, naming the resulting dataset as CC3M-Filtered. \textcolor{black}{We also apply this method to PraCegoVer\footnote{PraCegoVer filtered version: \scriptsize{\url{https://zenodo.org/records/7548638}}.}, used as a test set, to remove unrelated image-text pairs}.

\subsection{Dataset Re-annotation \& Translation}
\label{sec:dataset_reannotation}
CLIP is a framework based on contrastive learning to train a multimodal model. In its pipeline, a large batch of image-text pairs $(x_I, x_T)$ is sampled at each training step. Then, the image and text features are extracted by the respective encoders $f_T$ and $f_I$ and are used to compute InfoNCE loss~\cite{oord2018representation} as follows:\vspace{-0.25cm} 
\begin{equation}
\label{eq.infonce}
\resizebox{0.89\columnwidth}{!}{$
L_{\text{InfoNCE}}(x, y) \!=\! - \displaystyle\sum_{i=1}^{B} \log \frac{\exp(\text{sim}(x^i, y^i)/\tau}{\displaystyle\sum_{j=1}^{B} \exp(\text{sim}(x^i, y^j))/\tau},
$}
\end{equation}
\vspace{-0.15cm}
\begin{equation}
\label{eq.clip_loss}
\resizebox{0.89\columnwidth}{!}{$
L_{\text{CLIP}} = L_{\text{InfoNCE}}(f_I(\text{aug}(x_I)), f_T(x_T)),
$}
\end{equation}
\noindent where $B$ is the batch size, $\tau$ is a learnable temperature to scale the logits, $\text{sim}(\cdot)$ and $\text{aug}(\cdot)$ stands for cosine similarity and augmentation operation, respectively.

In the original proposal, only images are augmented as indicated in Equation~\ref{eq.clip_loss}, which might limit the text guidance to the image encoder. \citet{laclip_fan2023improving} propose to use LLM to augment texts in addition to the image augmentation, as shown in Equation~\ref{eq.infonce_text_augment}. However, this text-generation process does not consider the image content. 
\vspace{-0.1cm}
\begin{equation}
\label{eq.infonce_text_augment}
\resizebox{0.89\columnwidth}{!}{$
L_{\text{text aug.}} \!\!=\!\! L_{\text{InfoNCE}}(f_I(\text{aug}(x_I)), f_T(\text{aug}(x_T))).
$}
\end{equation}

We propose to use BLIP2\footnote{\scriptsize{\url{https://huggingface.co/Salesforce/blip2-opt-2.7b}}} to generate new captions conditioned on the images from CC3M. In contrast to~\citet{nguyen2023improving}, and drawing inspiration from LaCLIP~\cite{laclip_fan2023improving}, we propose to generate multiple captions for each image in the dataset by passing different prefixes to BLIP2. Our approach addresses the limitation of LaCLIP and has the advantage of generating multiple captions per image, which is a drawback of~\citet{nguyen2023improving}. Still, as BLIP2 
\textcolor{black}{is a monolingual model}, we decided to generate the captions in English and then translate them into Portuguese using Google Translate\footnote{\scriptsize\url{https://translate.google.com.br}}. Therefore, our text augmentation comprehends generating English captions with BLIP2 and translating them into Portuguese. During training, for each image, we randomly sample a caption among the original and the generated ones to fine-tune the text encoder. Hence, at each epoch, a different text can be selected for each image. For evaluation, we translate the annotations from Flickr30k and MS COCO, and the labels from ImageNet and ELEVATER.

\subsection{Training}
\label{sec:training}
This work takes place within the context of limited computational resources. We apply many techniques to reduce the cost of fine-tuning the \textsc{OpenCLIP}. First, we use Gradient Checkpointing~\cite{chen2016training}, which reduces the memory usage to $O(\sqrt{n})$ when training $n$ layers. This method removes the layers' activation after the forward pass and recalculates them during the backward pass if necessary. Using this technique, we achieved a considerable reduction in GPU memory usage.

Another method contributing to memory reduction is LiT~\cite{lit_Zhai_2022}, which only trains the text encoder while keeping the image encoder frozen. The motivation for training only the text encoder is that the image encoder has already undergone extensive pre-training and can produce good representations for images. Hence, we train the text encoder with captions in Portuguese so that this model learns to align the text embeddings to fixed image features, producing a multimodal embedding space. This strategy speeds up training and reduces memory since the image encoder does not compute gradients.

Finally, we also apply LoRA~\cite{hu2021lora} to reduce the number of trainable parameters, reducing the memory needed to train the models and the training time. LoRA involves a re-parameterization of the dense layers as follows:\vspace{-0.2cm}
\begin{equation}
    h = W_ox + \frac{\alpha}{r}BAx,\vspace{-0.2cm}
\end{equation}
\noindent where $W_o \in \mathbb{R}^{d_1\times d_2}$ is the frozen pre-trained weight matrix, $h$ is the result of the re-parameterization, $A \in \mathbb{R}^{r\times d_2}$ and $B \in \mathbb{R}^{d_1\times r}$ are decomposition matrices and $r < \min(d_1, d_2)$ is the low-dimensional rank of the decomposition, an $\alpha$ is a hyperparameter for scale. \textcolor{black}{Similar to \citet{hu2021lora}}, we use LoRA in the query (Q) and \textcolor{black}{value~(V)} self-attention modules from the text encoder.

The original \textsc{OpenCLIP} consists of 366M parameters. Applying LiT strategies reduces this number to 88M trainable parameters (24\% of the total). Further integration of LoRA reduces the trainable parameters to only 0.1\% (300k). We report all the training hyperparameters in~Appendix~\ref{sec:app-hyperparamters}.

\subsection{Evaluation}
\label{sec:evaluation}
To evaluate the proposed framework's generalization capacity, we follow the typical procedure of evaluating pre-trained models~\cite{radford2021learning, cn_clip_yang_2022, kelip_Ko_Gu_2022} in zero-shot cross-modal retrieval (text-to-image and image-to-text retrieval) and zero-shot image classification. 

\textbf{Zero-shot Cross-modal Retrieval}: We evaluate our methods on three cross-modal retrieval datasets: PraCegoVer, MS COCO, and Flickr30k. PraCegoVer is a multimodal dataset with native Portuguese captions based on Instagram posts. We built upon the conventional MS COCO and Flickr30k datasets, using Google Translate to translate all captions to Portuguese. To assess cross-modal retrieval performance, we adopted the recall@$K$ evaluation metric, where $K=\{1, 5, 10\}$, and the mean recall, representing the average recall value across the recall@$K$ instances.

\textbf{Zero-shot Image Classification}: We evaluate our pre-trained models on ImageNet-1k~\cite{deng2009imagenet} and on ELEVATER image classification toolkit~\cite{li2022elevater}. It contains 20 datasets designed for image classification tasks across various domains and an easy-to-use toolkit to evaluate pre-trained language-augmented visual models. To accommodate evaluation in the Portuguese language, we manually translated the labels for each dataset, as well as the templates, following the methodology outlined in~\cite{radford2021learning}. In the evaluation process, ImageNet-1k employs the top-1 accuracy metric. Appendix~\ref{sec:results_elevater_imagenet} provides the specific metrics for each dataset in ELEVATER~benchmark.

\section{Experiments and Results}
\label{sec:experiments}
This section presents a comprehensive set of experiments designed to investigate the effects of dataset filtering and the specific influence of each module within our framework, CAPIVARA. To reduce the effects of randomness, we ran each experiment setup three times. We also focus on zero-shot tasks involving images and Portuguese texts. Since no CLIP model is publicly available for Portuguese, we adopt as baseline the pre-trained multilingual model \textsc{OpenCLIP} due to its state-of-the-art performance in many tasks with Portuguese captions.

\vspace{-0.05cm}\paragraph{Dataset Filtering \& CAPIVARA.}\hspace{-0.15cm}We investigate two data-centric approaches: filter the training set by selecting promising samples capable of removing noise, and annotation enhancement with our proposed framework. Using CAPIVARA, for each image in CC3M, we add 10 synthetic captions, generated with BLIP2, besides the original caption. We comprehensively analyze the impact of the dataset filtering presented in Sec.~\ref{sec:dataset_filtering} and the effectiveness of CAPIVARA in cross-modal retrieval tasks on Flickr30k, MS COCO (with Portuguese-translated captions), and PraCegoVer datasets.

Table~\ref{table:re-annotation-translation} shows the results of the text-to-image (txt2img) and image-to-text (img2txt) retrieval tasks conducted on \textsc{OpenCLIP}. These results encompass models fine-tuned and trained using the CAPIVARA framework on the original CC3M dataset and its filtered version, CC3M-Filtered. In Table~\ref{table:re-annotation-translation}, the columns ``Synth.'' and ``Trans.'' indicate which settings include synthetic captions and whether or not the captions are translated.

\begin{table*}[t]
\setlength{\tabcolsep}{1.8mm}
\centering
\caption{
Impact analysis of synthetic captions (Synth.) and translation (Trans.) on our framework. This table compares the performance of CLIP fine-tuning on English and Portuguese-translated texts, both with and without the addition of synthetic captions. It shows the experimental results in cross-modal retrieval on Flickr30k and MS COCO with captions translated into Portuguese, and PraCegoVer. We report the average and standard deviation of mean recall for text-to-image (txt2img) and image-to-text (img2txt) retrieval tasks. Our CAPIVARA achieves the best performance across datasets, highlighting its efficacy in enhancing pre-trained multilingual CLIP.}\vspace{-0.2cm}

\label{table:re-annotation-translation}
\scriptsize
\begin{tabular}{llccllllll}
\cline{5-10}
 &  & & & \multicolumn{2}{c}{Flickr30k} & \multicolumn{2}{c}{MS COCO} & \multicolumn{2}{c}{PraCegoVer} \\ \hline
 Method/Model  & Training dataset & Synth. & Trans. & txt2img  & img2txt & txt2img & img2txt & txt2img & img2txt\\ \hline
 \textsc{OpenCLIP} (Baseline) & & & & 76.23 & 87.93 & 52.62 & 66.55 & 65.36	 & \textbf{69.43} \\ \hline
\multicolumn{1}{l}{\multirow{4}{*}{\begin{tabular}[l]{@{}l@{}}\textsc{OpenCLIP}\\+ Fine-tuning\end{tabular}}}  & CC3M &   \xMark & \xMark &
75.78 {\tiny ± 0.02} & 
88.78 {\tiny ± 0.04} & 
52.28 {\tiny ± 0.01} & 
68.18 {\tiny ± 0.01} & 
61.41 {\tiny ± 0.00} &	
62.35 {\tiny ± 0.01}\\ 
 & CC3M & \checkMark & \xMark &
77.08 {\tiny ± 0.02} & 
89.01 {\tiny ± 0.03} & 
53.87 {\tiny ± 0.01} & 
70.04 {\tiny ± 0.02} & 
64.01 {\tiny ± 0.01} &	
66.43 {\tiny ± 0.01} \\
& CC3M & \xMark & \checkMark &
78.42 {\tiny ± 0.02}  &
\textbf{90.02 {\tiny ± 0.05}}  & 
54.77 {\tiny ± 0.01}  &
70.06 {\tiny ± 0.01}  &
63.79 {\tiny ± 0.01} & 60.10 {\tiny ± 0.00} \\
 & CC3M-Filtered &  \xMark & \checkMark &
79.02 {\tiny ± 0.01} & 
89.49 {\tiny ± 0.02} & 
55.46 {\tiny ± 0.01} & 
69.52 {\tiny ± 0.02} & 
65.11 {\tiny ± 0.01} & 
62.29 {\tiny ± 0.01} \\
 \hline
\multirow{2}{*}{\capivaraicon{1.5em} \textbf{CAPIVARA}} & CC3M & \checkMark & \checkMark &
\textbf{79.56 {\tiny ± 0.01}} & 
\textbf{89.95 {\tiny ± 0.04}} & 
\textbf{56.27 {\tiny ± 0.01}} & 
\textbf{71.24 {\tiny ± 0.01}} & 
\textbf{66.40	{\tiny ± 0.01}} & 
64.75 {\tiny ± 0.01} \\
 & CC3M-Filtered & \checkMark & \checkMark &
 \textbf{79.67 {\tiny ± 0.01}}  & 
 \textbf{89.97 {\tiny ± 0.04}}  & 
 \textbf{56.32 {\tiny ± 0.01}}  & 
 \textbf{71.06 {\tiny ± 0.01}}  &
 \textbf{66.55 {\tiny ± 0.01}}  &	
 65.06 {\tiny ± 0.01} \\ \hline
\end{tabular}
\end{table*}

Employing the CC3M \textcolor{black}{with translated captions, fourth row in Table~\ref{table:re-annotation-translation},} for fine-tuning increases the mean recall score by roughly 2~percentage points (\textit{pp.})~in text-to-image and image-to-text retrieval tasks on Flickr30k and MS COCO, compared to the baseline, \textsc{OpenCLIP}. However, for the PraCegoVer dataset, a decline of 1.6~\textit{pp.}~in text-to-image retrieval and a more significant drop of 9.3~\textit{pp.}~in image-to-text retrieval are observed. Comparing the fine-tuning using CC3M and CC3M-Filtered, one can note an average enhancement of 0.9 \textit{pp.}~in mean recall score for text-to-image retrieval and a 0.4 \textit{pp.}~improvement for image-to-text retrieval across all three datasets.

\textcolor{black}{In addition, as an intermediate step in our architecture, we employ synthetic captions to mitigate noise in the training data. To illustrate the performance gains, we compare the results of only translating the training set and translating and generating synthetic captions (CAPIVARA), fourth and sixth rows in Table~\ref{table:re-annotation-translation}, respectively. For the Flickr30k dataset, we observe a 1.1~\textit{pp.}~improvement in text-to-image retrieval with synthetic captions, with no significant difference in image-to-text retrieval. On the MS COCO dataset, we note a 1.5~\textit{pp.}~increase in text-to-image retrieval and a 1.2~\textit{pp.}~gain in image-to-text retrieval. Additionally, when evaluating the PraCegoVer dataset under the same conditions, we find a 2.6~\textit{pp.}~improvement in text-to-image retrieval and a 4.7~\textit{pp.}~gain in image-to-text retrieval. Thus, in most cases, using synthetic data as a means of data augmentation and noise reduction yields a positive impact. Details about the impact of the number of synthetic captions in the performance are shown in Table \ref{tab:multiple-captions} (Appendix~\ref{sec:ablation_study}).
}

The most significant performance gains over the baseline are achieved using CAPIVARA. For instance, the model trained on CC3M with CAPIVARA, sixth row, yields a 3.5~\textit{pp.}~improvement in text-to-image retrieval for Flickr30k and MS COCO and 1~\textit{pp.}~enhancement on PraCegoVer.  Notably, in image-to-text retrieval, CAPIVARA (CC3M) increases 2~\textit{pp.}~on Flickr30k and it has a remarkable 4.7 \textit{pp.}~gain on MS COCO over the baseline. Also, models trained on CC3M and CC3M-Filtered with CAPIVARA demonstrate similar performance levels. These experiments demonstrate the effectiveness of our proposal, CAPIVARA, in enhancing multilingual CLIP performance in Portuguese.

\vspace{-0.1cm}\paragraph{Caption Translation.}\hspace{-0.15cm}We also investigate the impacts of automatic translations of captions in the final model performance for Portuguese texts. We conducted experiments training the model on datasets containing only English annotations (i.e., CC3M + no-translation and CC3M + no-translation + synthetic captions), and their counterparts translated into Portuguese using Google Translate (i.e., CC3M + translation and CC3M + translation + synthetic captions). The evaluation comprehends Flickr30k, MS COCO, and PraCegoVer datasets with only Portuguese captions, particularly images in PraCegoVer that are originally annotated in Portuguese. We present the results in Table~\ref{table:re-annotation-translation}.

One can note a substantial improvement when translating annotations within the training dataset. Specifically, models trained on datasets containing Portuguese annotations exhibit an average increase of 2.5 \textit{pp.}~in text-to-image mean recall scores compared to their English-trained counterparts. Similarly, employing Portuguese-translated captions leads to a mean recall improvement of 1.6 \textit{pp.}~for image-to-text retrieval on both the Flickr30k and MS COCO datasets. Fine-tuning with the original CC3M (i.e., CC3M + no-translation) hampers text-to-image performance across all three datasets and drops notable 7 \textit{pp.}~the mean recall in image-to-text on PraCegoVer. By training the model on translated synthetic captions, CAPIVARA consistently outperformed all the other settings. Our method increases the average performance in 3.2 \textit{pp.}~compared to fine-tuning on the original CC3M dataset. This experiment highlights the importance of including the automatic translation of captions into the target language, Portuguese, in our training pipeline.

\vspace{-0.1cm}\paragraph{Training Pipeline Optimization.}\hspace{-0.15cm}\label{sec:pipeline_optimization}This work is inserted in a context of restricted computational resources, in which only a single RTX Quadro 8000 GPU is available. In this way, we propose a method to optimize our training pipeline, detailed in Sec.~\ref{sec:training}. It combines LiT, Gradient Checkpointing (G. Checkpt), and LoRA techniques. In this section, we investigate the impacts of this optimization in terms of model performance and cost reduction. All experiments include LiT and gradient checkpointing, otherwise, we could not run the training in our infrastructure. In addition, we conducted experiments to assess the impact of including LoRA in our training pipeline. To compare the computational cost among the settings, we fixed the GPU architecture, and we trained the models with batch size (BS) equal to 2816 for 5863 steps, except for LiT + G. Checkpt + LoRA + 1500 steps + BS=1000, trained with a batch size of 1000 samples for only 1500 steps. Still, we demonstrate that it is possible to reduce the batch size and the number of training steps and reach a competitive~performance. 

\begin{table}[t]
\centering
\caption{Impact of optimization techniques. We evaluate training models on CC3M with CAPIVARA combined with many optimization techniques. We report the experimental results in terms of mean recall in text-to-image (txt2img), and image-to-text (img2txt) and memory (M) and training time cost (T). Our optimization method leads to the best training time and computational cost while performing similarly to other approaches.}\vspace{-0.2cm}
\label{table:otimization-results}
\resizebox{\columnwidth}{!}{%
\begin{tabular}{p{0.01cm}lcccccccc}
\cline{3-10}
 &
  \multicolumn{1}{c}{} &
  \multicolumn{2}{c}{Flickr30k} &
  \multicolumn{2}{c}{MS COCO} &
  \multicolumn{2}{c}{PraCegoVer} &
  \multirow{2}{*}{\begin{tabular}[c]{@{}c@{}}M\vspace{-0.15cm}\\ (GB)\end{tabular}} &
  \multirow{2}{*}{\begin{tabular}[c]{@{}c@{}}T\vspace{-0.15cm}\\ (h)\end{tabular}} \\ \cline{1-8}
 &
  \multicolumn{1}{c}{Optimization} &
  txt2img &
  img2txt &
  txt2img &
  img2txt &
  txt2img &
  img2txt &
   &
   \\ \cline{2-10} 
 & 
  \multicolumn{1}{l}{\begin{tabular}[l]{@{}l@{}}\textsc{OpenCLIP}\vspace{-0.1cm}\\(Baseline)\end{tabular}} &
  76.23 &
  87.93 &
  52.62 &
  66.55 &
  65.36 &
  69.43 &
  - &
  - \\ \cline{1-10} 
\multicolumn{1}{c}{\multirow{3}{*}{\rule{0pt}{17ex}\rotatebox{90}{CAPIVARA}}} &
  \begin{tabular}[c]{@{}l@{}}LiT + \\ G.Checkpt\end{tabular} &
  \begin{tabular}[c]{@{}l@{}}79.56\vspace{-0.2cm}\\{\tiny ± 0.01}\end{tabular} &
  \begin{tabular}[c]{@{}l@{}}89.95\vspace{-0.2cm}\\{\tiny ± 0.04}\end{tabular} &
  \begin{tabular}[c]{@{}l@{}}56.27\vspace{-0.2cm}\\{\tiny ± 0.01}\end{tabular} &
  \begin{tabular}[c]{@{}l@{}}71.24\vspace{-0.2cm}\\{\tiny ± 0.01}\end{tabular} &
  \begin{tabular}[c]{@{}l@{}}66.44\vspace{-0.2cm}\\{\tiny ± 0.01}\end{tabular} &
  \begin{tabular}[c]{@{}l@{}}66.57\vspace{-0.2cm}\\{\tiny ± 0.01}\end{tabular} &
  38 &
  31 \\ \cline{2-10} 
\multicolumn{1}{l}{} &
  \begin{tabular}[c]{@{}l@{}}LiT +\\ G.Checkpt +\\ LoRA\end{tabular} &
  \begin{tabular}[c]{@{}l@{}}79.51\vspace{-0.2cm}\\{\tiny ± 0.04}\end{tabular} &
  \begin{tabular}[c]{@{}l@{}}89.50\vspace{-0.2cm}\\{\tiny ± 0.03}\end{tabular} &
  \begin{tabular}[c]{@{}l@{}}55.56\vspace{-0.2cm}\\{\tiny ± 0.01}\end{tabular} &
  \begin{tabular}[c]{@{}l@{}}69.63\vspace{-0.2cm}\\{\tiny ± 0.04}\end{tabular} &
  \begin{tabular}[c]{@{}l@{}}67.07\vspace{-0.2cm}\\{\tiny ± 0.02}\end{tabular} &
  \begin{tabular}[c]{@{}l@{}}68.14\vspace{-0.2cm}\\{\tiny ± 0.01}\end{tabular} &
  21.5 &
  16 \\ \cline{2-10} 
\multicolumn{1}{l}{} &
  \begin{tabular}[c]{@{}l@{}}LiT +\\ G.Checkpt +\\ LoRA +\\ 1500 steps +\\ BS=1000\end{tabular} &
  \begin{tabular}[c]{@{}l@{}}79.39\vspace{-0.2cm}\\{\tiny ± 0.05}\end{tabular} &
  \begin{tabular}[c]{@{}l@{}}89.13\vspace{-0.2cm}\\{\tiny ± 0.08}\end{tabular} &
  \begin{tabular}[c]{@{}l@{}}55.49\vspace{-0.2cm}\\{\tiny ± 0.06}\end{tabular} &
  \begin{tabular}[c]{@{}l@{}}69.26\vspace{-0.2cm}\\{\tiny ± 0.05}\end{tabular} &
  \begin{tabular}[c]{@{}l@{}}66.89\vspace{-0.2cm}\\{\tiny ± 0.04}\end{tabular} &
  \begin{tabular}[c]{@{}l@{}}67.93\vspace{-0.2cm}\\{\tiny ± 0.01}\end{tabular} &
  8.5 &
  2 \\ \cline{1-10} 
\end{tabular}%
}
\end{table}

Table~\ref{table:otimization-results} shows experimental results. Our initial attempt to fine-tune the complete CLIP model encountered infrastructure limitations, hindering its execution. We overcame this constraint by utilizing LiT and gradient checkpointing, which enabled the training process. Comparison between the setups, namely LiT + G. Checkpt and LiT + G. Checkpt + LoRA, reveals that LoRA substantially reduces memory usage by over 40\% and cuts training time in half. The model trained with LoRA had a performance similar to the one that fine-tunes the entire text encoder on Flickr30k, but it decreases by 1.2~\textit{pp.}~the average performance on MS COCO.
 
In addition, the model trained with our optimization technique LiT + G. Checkpt + LoRA + 1500 steps + BS=1000 presented a decline of 0.2 \textit{pp.}~compared to LiT + G. Checkpt + LoRA. Using~our optimization method can remarkably reduce the GPU memory (from 38 GB to 8.5 GB) and training time (from 31h to 2h), yet outperform the baseline by 2.5 \textit{pp.}~across the tasks. Our training pipeline requires very modest computational resources compared to the literature, as shown in Table~\ref{tab:model-resources}. These experiments demonstrate that our optimization method can effectively reduce the~cost of fine-tuning CLIP, allowing researchers with restric\-ted computing resources to conduct experiments.

\begin{table}[t]
\centering
\caption{Summary of the models and resources invested in their training, considering the dataset size, the GPU/TPU used, and the required training time.}\vspace{-0.2cm}
\label{tab:model-resources}
\resizebox{\columnwidth}{!}{%
\begin{tabular}{lrrrr}
\hline
Model           & Language            & \# Dataset size & GPU/TPU                       & Training time         \\
\hline
Italian CLIP     & Italian             & 1.4M          & 2 TPUs                      & 14 days      \\
Chinese CLIP         & Chinese             & 200M          & 128 V100 (2048 GB)          & 7.5 days     \\
Korean CLIP           & Korean              & 708M          & 80 A100 (640 GB)            & 15.7 days    \\
LaCLIP          & English             & 365M          & 32 V100 (512 GB)            & --            \\
AltCLIP         & Multilingual        & 38M/115M      & --                           & --            \\
M-CLIP           & Multilingual        & 3.3M          & --                           & --            \\
\multirow{2}{*}{\textbf{CAPIVARA}} & \multirow{2}{*}{\textbf{Portuguese}}          & \multirow{2}{*}{\textbf{3.3M}}          & \textbf{1 Quadro RTX}   & \multirow{2}{*}{\textbf{2 hours}}      \\ 
&           &          & \textbf{8000 (48 GB})   &       \\ \hline
\end{tabular}%
}
\end{table}

\vspace{-0.1cm}\paragraph{Low-resource Languages.}\hspace{-0.15cm}
To demonstrate the effectiveness of CAPIVARA in improving pre-trained multilingual CLIP performance on low-resource languages, we expand our investigation to include Xhosa and Hindi. \textcolor{black}{Figure~\ref{fig:low-resource-performance} compares the performance between the pre-trained \textsc{OpenCLIP} (baseline) and the models trained by employing the whole CAPIVARA optimized pipeline, which refers to the setting LiT + G.~Checkpt + LoRA + 1500 steps + BS=1000, named CAPIVARA + Opt.,} for text-to-image and image-to-text retrieval tasks on Flickr30k and MS COCO. This experiment employs our optimized training pipeline (Sec.~\ref{sec:pipeline_optimization}), training models for 2 hours on a single GPU Quadro RTX 8000 with a memory usage of 8.5 GB.

The baseline presents the weakest performance in Xhosa across all tasks, with mean recall close to zero in MS COCO and 3 and 10 in text-to-image and image-to-text on Flickr30k, respectively. CAPIVARA increases the average performance in this language by 6.5 \textit{pp.}~on Flickr30k and MS COCO. The most significant improvement can be noted in Hindi. A remarkable increase of 15~\textit{pp.}~on MS COCO and 21 \textit{pp.}~on Flickr30k is obtained with CAPIVARA. This experiment shows that CAPIVARA effectively boosts the pre-trained multilingual CLIP's performance in other low-resource languages with a low computational cost.

\vspace{-0.05cm}\paragraph{Image Classification.}\hspace{-0.15cm}In addition to zero-shot cross-modal retrieval tasks, we also evaluate our models in zero-shot image classification across 21~datasets. The results are presented in Table~\ref{table:best-results-elevater}. In the context of ELEVATER, training CLIP with CAPIVARA yielded an average improvement of 0.6 \textit{pp.} over our baseline. We plot the bar chart in~Figure~\ref{fig:cc3m_10_gen} to thoroughly analyze the performance gap between the baseline and the model trained with CAPIVARA for each dataset within ELEVATER. Our method consistently surpassed the baseline across most datasets, yielding substantial accuracy improvements of 5.53 \textit{pp.}, 5.15 \textit{pp.}, and 3.07 \textit{pp.} for KITTI-Distance, MNIST, and GTSRB, respectively. Regarding ImageNet-1k, CAPIVARA exhibited a performance gain of 0.2 \textit{pp.} compared to the baseline. In addition, the model's performance trained with CAPIVARA + Opt.~is close to our baseline. Hence, LoRA-tuning for 1500 steps keeps the average performance on zero-shot image classification, whereas it improves considerably the performance on zero-shot cross-modal retrieval.

\begin{table}[t]
\centering
\caption{Zero-shot image classification performance on ELEVATER and ImageNet-1k.}\vspace{-0.2cm}
\label{table:best-results-elevater}
\resizebox{\columnwidth}{!}{%
\begin{tabular}{lccc}
\hline
Dataset &
  \multicolumn{1}{c}{\begin{tabular}[c]{@{}c@{}}\textsc{OpenCLIP}\vspace{-0.1cm}\\(Baseline)\end{tabular}} &
  CAPIVARA &
  CAPIVARA + Opt. \\ \hline
    Caltech-101        & 84.53 {\tiny ± 0.00} & 82.97 {\tiny ± 0.03} & 83.68 {\tiny ± 0.02} \\
    CIFAR-10           & 93.99 {\tiny ± 0.00} & 93.85 {\tiny ± 0.00} & 93.93 {\tiny ± 0.03} \\
    CIFAR-100          & 68.44 {\tiny ± 0.00} & 69.37 {\tiny ± 0.01} & 68.87 {\tiny ± 0.01} \\
    Country-211        & 17.82 {\tiny ± 0.00} & 17.61 {\tiny ± 0.00} & 17.32 {\tiny ± 0.02} \\
    DTD                & 41.17 {\tiny ± 0.00} & 42.34 {\tiny ± 0.04} & 41.79 {\tiny ± 0.07} \\
    EuroSAT            & 47.16 {\tiny ± 0.00} & 47.77 {\tiny ± 0.02} & 48.85 {\tiny ± 0.12} \\
    FER-2013           & 48.65 {\tiny ± 0.00} & 46.68 {\tiny ± 0.05} & 46.85 {\tiny ± 0.13} \\
    FGVC-Aircraft      & 26.30 {\tiny ± 0.00} & 25.49 {\tiny ± 0.01} & 25.54 {\tiny ± 0.09} \\
    Food-101           & 65.06 {\tiny ± 0.00} & 64.58 {\tiny ± 0.01} & 64.46 {\tiny ± 0.00} \\
    GTSRB              & 43.27 {\tiny ± 0.00} & 46.34 {\tiny ± 0.01} & 44.66 {\tiny ± 0.06} \\
    Hateful-Memes      & 56.50 {\tiny ± 0.00} & 56.17 {\tiny ± 0.00} & 56.81 {\tiny ± 0.03} \\
    KITTI-Distance     & 28.41 {\tiny ± 0.00} & 33.94 {\tiny ± 0.13} & 28.27 {\tiny ± 0.11} \\
    MNIST              & 54.99 {\tiny ± 0.00} & 60.14 {\tiny ± 0.04} & 55.00 {\tiny ± 0.10} \\
    Oxford Flowers-102 & 50.88 {\tiny ± 0.00} & 49.93 {\tiny ± 0.02} & 51.99 {\tiny ± 0.12} \\
    Oxford-IIIT Pets   & 81.56 {\tiny ± 0.00} & 79.37 {\tiny ± 0.00} & 80.90 {\tiny ± 0.09} \\
    PatchCamelyon      & 50.96 {\tiny ± 0.00} & 51.71 {\tiny ± 0.01} & 52.39 {\tiny ± 0.07} \\
    Rendered-SST2      & 54.20 {\tiny ± 0.00} & 54.82 {\tiny ± 0.03} & 52.94 {\tiny ± 0.04} \\
    RESISC-45          & 58.51 {\tiny ± 0.00} & 59.71 {\tiny ± 0.01} & 56.93 {\tiny ± 0.01} \\
    Stanford-Cars      & 84.93 {\tiny ± 0.00} & 85.10 {\tiny ± 0.02} & 84.90 {\tiny ± 0.06} \\
    PASCAL VOC-2007    & 82.09 {\tiny ± 0.00} & 82.29 {\tiny ± 0.00} & 81.99 {\tiny ± 0.02} \\ \hline
    Average            & 56.97 {\tiny ± 0.00} & \textbf{57.51 {\tiny ± 0.02}} & 56.90 {\tiny ± 0.06} \\ \hline
    ImageNet-1k         & 45.84 {\tiny ± 0.00} & \textbf{46.06 {\tiny ± 0.01}} & 45.65 {\tiny ± 0.02} \\ \hline
    \end{tabular}%
    }    
\end{table}

\vspace{-0.05cm}\paragraph{Carbon Footprint.}\hspace{-0.15cm}Despite the remarkable achievements of large language models, their deployment requires substantial computational power, resulting in significant energy usage. For instance, models such as GPT-3 and BLOOM consumed approximately 1,287 MWh and 433 MWh, respectively, in their training, corresponding to 502 tonnes of CO$_2$ and 25 tonnes of CO$_2$ emissions~\cite{ai_index_report}. The BLOOM model's carbon footprint alone surpasses an average American's annual carbon emissions by 1.4 times. The energy consumed during BLOOM's training could power a household in the United States for up to 41 years.

\begin{table}[t]
\centering
\caption{Average costs per trained model in terms of energy consumption and equivalent CO$_2$ emissions (\text{CO$_2$-eq}), compared with the number of trainable parameters (\# Param.). All the models were trained with a batch size (BS) of 2816 for 5863 steps, except for CAPIVARA + LoRA + 1500 steps / BS=1000.}\vspace{-0.2cm}
\label{tab:carbon-tracker}
\resizebox{\columnwidth}{!}{%
\begin{tabular}{lrlrlrl}
\hline
Model & \multicolumn{2}{c}{\hspace{-0.2cm}\# Param.} &
  \multicolumn{2}{c}{Energy} &
  \multicolumn{2}{c}{CO$_2$-eq} \\ \hline
\multicolumn{1}{l}{Gopher}   & 280 &\hspace{-0.3cm}B  & 1,066 &\hspace{-0.3cm}MWh & 352 &\hspace{-0.3cm}tonnes \\ 
\multicolumn{1}{l}{BLOOM}    & 176 &\hspace{-0.3cm}B  & 433 &\hspace{-0.3cm}MWh   & 25 &\hspace{-0.3cm}tonnes  \\ 
\multicolumn{1}{l}{GPT-3}    & 175 &\hspace{-0.3cm}B  & 1,287 &\hspace{-0.3cm}MWh & 502 &\hspace{-0.3cm}tonnes \\ 
\multicolumn{1}{l}{OPT}      & 175 &\hspace{-0.3cm}B  & 324 &\hspace{-0.3cm}MWh   & 70 &\hspace{-0.3cm}tonnes  \\ \hline
\multicolumn{1}{l}{CAPIVARA} & 278 &\hspace{-0.3cm}M & 6.49 &\hspace{-0.3cm}kW    & 0.50 &\hspace{-0.3cm}kg     \\ 
CAPIVARA + LoRA &
  1.9 &\hspace{-0.3cm}M &
  5.67 &\hspace{-0.3cm}kW &
  0.43 &\hspace{-0.3cm}kg \\ 
\multicolumn{1}{l}{\begin{tabular}[c]{@{}l@{}}CAPIVARA + LoRA\\ +1500 steps / BS=1000\end{tabular}} &
  1.9 &\hspace{-0.3cm}M &
  0.22 &\hspace{-0.3cm}kW &
  0.017 &\hspace{-0.3cm}kg \\ \hline
\end{tabular}%
}
\end{table}

To compare energy consumption between our model and larger language models, we employed the codecarbon tool~\cite{BenoitCourty_etal._2023}. The results are shown in Table~\ref{tab:carbon-tracker}. As other CLIP-like models do not provide energy and carbon expenditure data, we present a comparison with other large language models for which such data is available in the literature~\cite{ai_index_report}. For the baseline model, the energy usage amounted to 6.4 kW, resulting in 0.5 kg of CO$_2$ equivalent emissions. Applying LoRA and reducing the number of training steps decreased energy consumption to 5.6~kW and 1.8 kW, respectively, resulting in 0.4 kg and 0.1 kg of CO$_2$ equivalent emissions. These calculations are based on Brazil's energy mix, where hydropower is the primary energy source. This calculation does not include the carbon footprint of the initial pre-training performed by \textsc{OpenCLIP}, but only the training with CAPIVARA. We aim to advance sustainable AI systems development by employing these techniques and optimizing training times.

\section{Conclusion}
\label{sec:conclusion}
This work demonstrates the potential challenges of fine-tuning multilingual CLIP models within low-resource languages due to noisy annotations. To address this issue, we introduce \capivaraicon{1.35em} CAPIVARA, a cost-effective framework that leverages image captioning models to enhance the dataset annotations. We conducted extensive experiments involving dataset filtering, re-annotation, and automatic translation. CAPIVARA effectively boosts \textsc{OpenCLIP} performance for Portuguese texts, achieving state-of-the-art results in many zero-shot tasks. Our findings show the importance of dataset re-annotation and automatic translation.

We also propose optimizing our training pipeline using LiT, including LoRA and gradient checkpointing. Our results show a substantial improvement in Portuguese performance by fine-tuning the pre-trained \textsc{OpenCLIP} in a single GPU for 2 hours, and only 8.5 GB of memory --- considerably modest compared to literature. Moreover, we demonstrate that our framework is readily extensible to other low-resource languages. 

A direction for future research involves investigating the scalability of the proposed approach in terms of dataset and model size, building upon its success with base models. We also plan to explore different image captioning models and text decoding methods. Due to the cost of generating synthetic captions and translating them to Portuguese, there is interest in automating the process, possibly by improving BLIP2's performance in Portuguese. Besides, due to the success of LoRA, other parameter-efficient fine-tuning can be explored. Lastly, an interesting research question remains open: ``how many examples annotated in a low-resource language are necessary to achieve a performance comparable to English?''.



\section*{Limitations}
\label{sec:limitations}

\paragraph{Model.}\hspace{-0.15cm}
Unlike other studies that compare models with varying architectures and sizes~\cite{radford2021learning, cn_clip_yang_2022, li2021supervision, slip_mu_2021}, our research focuses on specific choices: the ViT-B/32 as image encoder and the XLM-Roberta Base as text encoder. Future work will explore different model sizes within our budget and consider alternative fine-tuning approaches, such as Parameter-Efficient Fine-Tuning (PEFT) \cite{liao2023peft}.

\paragraph{Data.}\hspace{-0.15cm}
Recent efforts to adapt CLIP for specific languages~\cite{kelip_Ko_Gu_2022, cn_clip_yang_2022, italianCLIP_bianchi_2021} have typically used datasets much larger than our study. Investigating scalability using training datasets could reveal the optimal trade-off between cost and performance.

Generating captions in languages such as Portuguese involves two steps: caption generation and machine translation; due to the lack of robust non-English image captioning models. Hence, future research could focus on fine-tuning image captioning models for target languages to streamline the process and improve accuracy. Our study used the BLIP2 model for caption generation, but exploring alternative models could enhance results. 

An additional limitation is the prevalent use of machine-translated datasets in various multilingual datasets~\cite{mCLIP_Carlsson_2022, mural_jain_2021, cn_clip_yang_2022, italianCLIP_bianchi_2021}. However, these datasets may not effectively capture unique expressions, cultural nuances, and proper nouns, leading to bias over-amplification, where biases from the source text become exaggerated in the translated output~\cite{hovy2021five, prabhumoye2020case, hovy2020you}.

\section*{Ethics Statement}

CAPIVARA is a cost-efficient framework designed to enhance the performance of multilingual CLIP models in low-resource languages. For this purpose, CAPIVARA augments text data using image captioning and machine translation to generate multiple synthetic captions in low-resource languages, and the training pipeline is optimized with LiT, LoRA, and gradient checkpointing to alleviate the computational cost. Intended to be used for general tasks, the model learns to represent in a joint space texts and images. It can be employed in text-to-image, image-to-text retrieval, and image classification tasks. The developed model is particularly intended for scientific researchers.

Based on known problems with image and language models, the model may present lower performance for under-represented and minority groups \cite{bender2021dangers}. To adapt the model to low-resource languages, we use texts translated from English; thus, the model does not represent the cultural and local aspects of the countries that speak these target languages. This can lead to linguistic biases and a lack of representativeness for the target groups.  

The datasets used comprehend texts from the~internet and carry biases; thus, the model may perform differently for data collected from other sources. Also, the datasets may contain~data with cultural, political, or religious positioning.  

Furthermore, CAPIVARA does not generate any type of data that could pose a risk to human life. However, our model can be adapted for other specific tasks, e.g., image or text generation, which could contribute to generating false information and harming people. CAPIVARA is a framework that aims to improve performance for low-resource languages. However, our results show that despite the significant improvements achieved with CAPIVARA, there is still a considerable gap between the model performance with English texts and texts in low-resource languages. Further research is needed to improve performance across languages and incorporate cultural and linguistic elements into the~model.

Since language models require large computational, environmental, and financial resources~\cite{bender2021dangers}, CAPIVARA optimizes its training pipeline, resulting in a smaller carbon footprint than traditional fine-tuning. More details about ethical considerations can be found in Model Cards (Appendix~\ref{sec:model_cards}).

\section*{Acknowledgements}
This project was supported by the Ministry of Science, Technology, and Innovation of Brazil, with resources granted by the Federal Law 8.248 of October 23, 1991, under the PPI-Softex. The project was coordinated by Softex and published as Intelligent agents for mobile platforms based on Cognitive Architecture technology \text{[01245.013778/2020-21]}.

D.A.B.M. is partially funded by FAPESP 2023/05939-5.
A.I.F., T.S., N.S. are partially funded by Centro de Excel\^encia em Intelig\^encia Artificial (CEIA), da Universidade Federal de Goi\'as (UFG). E.L.C. is partially funded by CNPq 315468/2021-1. H.P. is partially funded by CNPq 304836/2022-2. S.A. is partially funded by CNPq 315231/2020-3, FAPESP 2013/08293-7, 2020/09838-0, Google Award for Inclusion Research 2022.

\bibliography{custom}
\bibliographystyle{acl_natbib}

\appendix

\section{Appendix}
\label{sec:appendix}

\setcounter{table}{0}
\renewcommand{\thetable}{A\arabic{table}}

\setcounter{figure}{0}
\renewcommand{\thefigure}{A\arabic{figure}}

\subsection*{Authors’ Contributions}

G.O.S., D.A.B.M., and A.I.F. collaborated on dataset translation, designing and implementing the proposed pipeline, analyzing the results, and writing the manuscript. G.O.S. also conducted experiments related to dataset filtering, re-annotation, translation, and low-resource languages. D.A.B.M. worked on constructing training datasets, focused on experiments to optimize the pipeline and conducted a carbon footprint analysis. A.I.F. executed inferences for zero-shot image classification. In collaboration with G.O.S., D.A.B.M., and A.I.F., J.S. wrote the Ethics Statement and Model Cards sections. L.P. helped in the result analysis. P.B. contributed to dataset translation. T.S. helped in constructing training datasets.  H.M. contributed to the discussion with the team and the result analysis. N.S. advised A.I.F. and T.S. throughout all tasks. E.C. advised G.O.S. throughout all tasks. H.P. advised the team on all tasks and contributed to the writing process. S.A. served as the principal advisor of the team, providing guidance on all tasks and contributing to the writing process. All authors reviewed the manuscript and provided critical feedback to enhance its quality.

\subsection{Hyperparameters}
\label{sec:app-hyperparamters}

To facilitate the reproducibility of the work, we present Tables \ref{tab:hyperparameters} and \ref{tab:lora-hyperparameters}. These tables contain the hyperparameters used for the best models evaluated in the different experiments. Table~\ref{tab:hyperparameters} contains only the hyperparameters used in the fine-tuning of the \textsc{OpenCLIP} model for Portuguese. Table~\ref{tab:lora-hyperparameters} considers the hyperparameters with the LoRA-tuning for the models with optimizations and 1500 steps, in Portuguese, Hindi and Xhosa.

\begin{table}[h]
\centering
\caption{Hyperparameters used in the fine-tuning.}
\label{tab:hyperparameters}\vspace{-0.2cm}
\small
\begin{tabular}{ll}
\hline
Hyperparameters        & Value                     \\ \hline
Batch size             & 2816                      \\
Maximum token length   & 77                        \\
Optimizer              & Adam                      \\
Weight decay           & 0.2                       \\
Adam $\epsilon$        & 1e-8                      \\
Adam $\beta$           & {[}0.9, 0.98{]}           \\
Learning rate schedule & CosineWarmupLR            \\
Maximum learning rate  & 5e-7                      \\
Minimum learning rate  & 1e-7                      \\
\# Steps               & 5863                      \\\hline
\end{tabular}
\end{table}

\begin{table}[h]
\centering
\caption{Hyperparameters used in LoRA-tuning.}\vspace{-0.2cm}
\label{tab:lora-hyperparameters}
\small
\begin{tabular}{ll}
\hline
Hyperparameters & Value          \\ \hline
LoRA r           & 8              \\
LoRA Alpha       & 8              \\
LoRA dropout     & 0              \\
bias             & None           \\
Target modules   & (query, value) \\
Modules to save  & projection     \\ \hline
Batch size             & 1000                    \\
Maximum token length   & 77                      \\
Optimizer              & Adam                    \\
Weight decay           & 0.2                     \\
Adam $\epsilon$        & 1e-8                    \\
Adam $\beta$           & {[}0.9, 0.98{]}         \\
Learning rate schedule & CosineWarmupLR          \\
Maximum learning rate  & 1e-5                    \\
Minimum learning rate  & 1e-6                    \\
\# Steps               & 1500                    \\\hline
\end{tabular}%
\end{table}

\subsection{Results on ELEVATER and ImageNet-1k}
\label{sec:results_elevater_imagenet}

In our supplementary experiments on ELEVATER and ImageNet-1k benchmarks, summarized in Table~\ref{tab:full-results-classification}, we consistently observe that our approach outperforms the baseline model across various setups, with the exception of CAPIVARA + Opt. This suggests that more training steps might be necessary to fully leverage LoRA's potential in fine-tuning. Furthermore, Table~\ref{tab:full-results-classification} reveals the effect of caption generation and filtering on the efficacy of our method. By analyzing the scenarios with synthetic captions, one can note that training with multiple captions per image outperforms training on only \textsc{OpenCLIP} + Fine-tuning both with or without filtering. Notably, the optimal configuration involves training with CAPIVARA on CC3M-Filtered, resulting in a performance boost of 0.6 \textit{pp.} over the baseline. Still, similar to the cross-modal retrieval in Sec.~\ref{sec:caption_selection},  we do not observe a significant performance gain by augmenting the number of generated captions. Table~\ref{tab:elevaterdatasets} provides the specific metrics for each dataset in ELEVATER benchmark.

\begin{table*}[t]
\caption{Results on ELEVATER benchmark. Ablation without LoRA and with LoRA.}\vspace{-0.2cm}
\label{tab:full-results-classification}
\centering
\resizebox{\textwidth}{!}{%
\begin{tabular}{lccccccccccc}
\hline
Dataset &
  \begin{tabular}[c]{@{}c@{}}\textsc{OpenCLIP}\vspace{-0.1cm}\\(Baseline)\end{tabular} &
  \begin{tabular}[c]{@{}c@{}}\textsc{OpenCLIP}\\ + Fine-tuning\end{tabular} &
  \begin{tabular}[c]{@{}c@{}}\textsc{OpenCLIP}\\ + Fine-tuning\\  (CC3M-Filtered)\end{tabular} &
   \begin{tabular}[c]{@{}c@{}}CAPIVARA\\ (CC3M-Filtered)\end{tabular} &
  \begin{tabular}[c]{@{}c@{}}CAPIVARA\end{tabular} &
  \begin{tabular}[c]{@{}c@{}}CAPIVARA\\ + 5 synth. \\captions\end{tabular} &
  \begin{tabular}[c]{@{}c@{}}CAPIVARA\\ + 1 synth.\\caption\end{tabular} &
  \begin{tabular}[c]{@{}c@{}}\textsc{OpenCLIP}\\ + Fine-tuning\\ + LoRA\end{tabular} &
  \begin{tabular}[c]{@{}c@{}}CAPIVARA\\ + LoRA \end{tabular} &
  \begin{tabular}[c]{@{}c@{}}\textsc{CAPIVARA}\vspace{-0.1cm}\\ + Opt.\end{tabular} \\ \hline
Caltech-101 &
  84.53 {\scriptsize ± 0.00} &
  82.50 {\scriptsize ± 0.01} &
  82.23 {\scriptsize ± 0.01} &
  82.90 {\scriptsize ± 0.00} &
  82.97 {\scriptsize ± 0.03} &
  82.66 {\scriptsize ± 0.00} &
  82.87 {\scriptsize ± 0.01} &
  83.06 {\scriptsize ± 0.07} &
  83.70 {\scriptsize ± 0.01} &
  83.68 {\scriptsize ± 0.02} \\
CIFAR-10 &
  93.99 {\scriptsize ± 0.00} &
  94.10 {\scriptsize ± 0.00} &
  93.93 {\scriptsize ± 0.00} &
  93.94 {\scriptsize ± 0.00} &
  93.85 {\scriptsize ± 0.00} &
  93.87 {\scriptsize ± 0.00} &
  93.96 {\scriptsize ± 0.00} &
  94.05 {\scriptsize ± 0.01} &
  93.96 {\scriptsize ± 0.01} &
  93.93 {\scriptsize ± 0.03} \\
CIFAR-100 &
  68.44 {\scriptsize ± 0.00} &
  69.13 {\scriptsize ± 0.01} &
  68.98 {\scriptsize ± 0.01} &
  69.33 {\scriptsize ± 0.01} &
  69.37 {\scriptsize ± 0.01} &
  69.37 {\scriptsize ± 0.01} &
  69.27 {\scriptsize ± 0.01} &
  69.07 {\scriptsize ± 0.00} &
  68.97 {\scriptsize ± 0.01} &
  68.87 {\scriptsize ± 0.01} \\
Country-211 &
  17.82 {\scriptsize ± 0.00} &
  17.80 {\scriptsize ± 0.01} &
  17.73 {\scriptsize ± 0.01} &
  17.63 {\scriptsize ± 0.01} &
  17.61 {\scriptsize ± 0.00} &
  17.79 {\scriptsize ± 0.00} &
  17.78 {\scriptsize ± 0.00} &
  17.63 {\scriptsize ± 0.00} &
  17.36 {\scriptsize ± 0.02} &
  17.32 {\scriptsize ± 0.02} \\
DTD &
  41.17 {\scriptsize ± 0.00} &
  42.36 {\scriptsize ± 0.03} &
  42.59 {\scriptsize ± 0.03} &
  42.59 {\scriptsize ± 0.05} &
  42.34 {\scriptsize ± 0.04} &
  42.62 {\scriptsize ± 0.03} &
  42.61 {\scriptsize ± 0.00} &
  41.52 {\scriptsize ± 0.05} &
  41.95 {\scriptsize ± 0.05} &
  41.79 {\scriptsize ± 0.07} \\
EuroSAT &
  47.16 {\scriptsize ± 0.00} &
  50.45 {\scriptsize ± 0.04} &
  50.51 {\scriptsize ± 0.02} &
  48.14 {\scriptsize ± 0.03} &
  47.77 {\scriptsize ± 0.02} &
  49.19 {\scriptsize ± 0.05} &
  50.03 {\scriptsize ± 0.03} &
  48.21 {\scriptsize ± 0.02} &
  48.53 {\scriptsize ± 0.08} &
  48.85 {\scriptsize ± 0.12} \\
FER-2013 &
  48.65 {\scriptsize ± 0.00} &
  46.08 {\scriptsize ± 0.03} &
  46.78 {\scriptsize ± 0.02} &
  46.93 {\scriptsize ± 0.03} &
  46.68 {\scriptsize ± 0.05} &
  46.80 {\scriptsize ± 0.01} &
  46.44 {\scriptsize ± 0.01} &
  47.93 {\scriptsize ± 0.01} &
  47.00 {\scriptsize ± 0.06} &
  46.85 {\scriptsize ± 0.13} \\
FGVC-Aircraft &
  26.30 {\scriptsize ± 0.00} &
  25.56 {\scriptsize ± 0.02} &
  25.70 {\scriptsize ± 0.01} &
  25.52 {\scriptsize ± 0.04} &
  25.49 {\scriptsize ± 0.01} &
  25.74 {\scriptsize ± 0.02} &
  25.70 {\scriptsize ± 0.01} &
  26.45 {\scriptsize ± 0.01} &
  26.23 {\scriptsize ± 0.03} &
  25.54 {\scriptsize ± 0.09} \\
Food-101 &
  65.06 {\scriptsize ± 0.00} &
  63.83 {\scriptsize ± 0.00} &
  64.27 {\scriptsize ± 0.01} &
  64.54 {\scriptsize ± 0.01} &
  64.58 {\scriptsize ± 0.01} &
  64.52 {\scriptsize ± 0.00} &
  64.21 {\scriptsize ± 0.02} &
  64.52 {\scriptsize ± 0.01} &
  64.67 {\scriptsize ± 0.00} &
  64.46 {\scriptsize ± 0.00} \\
GTSRB &
  43.27 {\scriptsize ± 0.00} &
  46.06 {\scriptsize ± 0.02} &
  46.95 {\scriptsize ± 0.01} &
  46.81 {\scriptsize ± 0.03} &
  46.34 {\scriptsize ± 0.01} &
  46.33 {\scriptsize ± 0.03} &
  46.62 {\scriptsize ± 0.02} &
  44.64 {\scriptsize ± 0.01} &
  44.88 {\scriptsize ± 0.06} &
  44.66 {\scriptsize ± 0.06} \\
Hateful-Memes &
  56.50 {\scriptsize ± 0.00} &
  56.06 {\scriptsize ± 0.01} &
  56.25 {\scriptsize ± 0.01} &
  56.09 {\scriptsize ± 0.01} &
  56.17 {\scriptsize ± 0.00} &
  55.98 {\scriptsize ± 0.01} &
  56.03 {\scriptsize ± 0.00} &
  57.01 {\scriptsize ± 0.01} &
  56.64 {\scriptsize ± 0.02} &
  56.81 {\scriptsize ± 0.03} \\
KITTI-Distance &
  28.41 {\scriptsize ± 0.00} &
  30.80 {\scriptsize ± 0.00} &
  30.24 {\scriptsize ± 0.11} &
  33.19 {\scriptsize ± 0.11} &
  33.94 {\scriptsize ± 0.13} &
  32.21 {\scriptsize ± 0.00} &
  29.96 {\scriptsize ± 0.00} &
  26.30 {\scriptsize ± 0.00} &
  28.36 {\scriptsize ± 0.07} &
  28.27 {\scriptsize ± 0.11} \\
MNIST &
  54.99 {\scriptsize ± 0.00} &
  53.64 {\scriptsize ± 0.04} &
  54.83 {\scriptsize ± 0.02} &
  61.86 {\scriptsize ± 0.02} &
  60.14 {\scriptsize ± 0.04} &
  59.57 {\scriptsize ± 0.01} &
  56.06 {\scriptsize ± 0.03} &
  55.68 {\scriptsize ± 0.04} &
  55.37 {\scriptsize ± 0.06} &
  55.00 {\scriptsize ± 0.10} \\
Oxford Flowers-102 &
  50.88 {\scriptsize ± 0.00} &
  49.98 {\scriptsize ± 0.00} &
  49.72 {\scriptsize ± 0.03} &
  49.74 {\scriptsize ± 0.02} &
  49.93 {\scriptsize ± 0.02} &
  50.03 {\scriptsize ± 0.02} &
  50.07 {\scriptsize ± 0.00} &
  51.26 {\scriptsize ± 0.01} &
  51.91 {\scriptsize ± 0.04} &
  51.99 {\scriptsize ± 0.12} \\
Oxford-IIIT Pets &
  81.56 {\scriptsize ± 0.00} &
  79.52 {\scriptsize ± 0.02} &
  80.69 {\scriptsize ± 0.01} &
  79.60 {\scriptsize ± 0.03} &
  79.37 {\scriptsize ± 0.00} &
  79.24 {\scriptsize ± 0.02} &
  79.46 {\scriptsize ± 0.01} &
  81.29 {\scriptsize ± 0.02} &
  81.24 {\scriptsize ± 0.03} &
  80.90 {\scriptsize ± 0.09} \\
PatchCamelyon &
  50.96 {\scriptsize ± 0.00} &
  57.15 {\scriptsize ± 0.01} &
  55.70 {\scriptsize ± 0.01} &
  51.93 {\scriptsize ± 0.00} &
  51.71 {\scriptsize ± 0.01} &
  52.56 {\scriptsize ± 0.03} &
  55.49 {\scriptsize ± 0.02} &
  52.86 {\scriptsize ± 0.02} &
  52.23 {\scriptsize ± 0.01} &
  52.39 {\scriptsize ± 0.07} \\
Rendered-SST2 &
  54.20 {\scriptsize ± 0.00} &
  53.05 {\scriptsize ± 0.04} &
  53.82 {\scriptsize ± 0.09} &
  53.67 {\scriptsize ± 0.03} &
  54.82 {\scriptsize ± 0.03} &
  54.35 {\scriptsize ± 0.03} &
  53.03 {\scriptsize ± 0.03} &
  53.47 {\scriptsize ± 0.03} &
  53.14 {\scriptsize ± 0.07} &
  52.94 {\scriptsize ± 0.04} \\
RESISC-45 &
  58.51 {\scriptsize ± 0.00} &
  58.78 {\scriptsize ± 0.01} &
  58.92 {\scriptsize ± 0.02} &
  59.56 {\scriptsize ± 0.01} &
  59.71 {\scriptsize ± 0.01} &
  59.25 {\scriptsize ± 0.02} &
  58.88 {\scriptsize ± 0.01} &
  57.06 {\scriptsize ± 0.00} &
  57.21 {\scriptsize ± 0.02} &
  56.93 {\scriptsize ± 0.01} \\
Stanford-Cars &
  84.93 {\scriptsize ± 0.00} &
  85.00 {\scriptsize ± 0.01} &
  85.04 {\scriptsize ± 0.01} &
  85.10 {\scriptsize ± 0.00} &
  85.10 {\scriptsize ± 0.02} &
  85.08 {\scriptsize ± 0.01} &
  85.08 {\scriptsize ± 0.01} &
  85.35 {\scriptsize ± 0.02} &
  84.99 {\scriptsize ± 0.03} &
  84.90 {\scriptsize ± 0.06} \\
PASCAL VOC-2007 &
  82.09 {\scriptsize ± 0.00} &
  82.73 {\scriptsize ± 0.00} &
  82.31 {\scriptsize ± 0.00} &
  82.24 {\scriptsize ± 0.01} &
  82.29 {\scriptsize ± 0.00} &
  82.39 {\scriptsize ± 0.00} &
  82.67 {\scriptsize ± 0.01} &
  82.35 {\scriptsize ± 0.00} &
  82.00 {\scriptsize ± 0.01} &
  81.99 {\scriptsize ± 0.02} \\ \hline
Average &
  56.97 {\scriptsize ± 0.00} &
  57.23 {\scriptsize ± 0.02} &
  57.36 {\scriptsize ± 0.02} &
  57.57 {\scriptsize ± 0.02} &
  57.51 {\scriptsize ± 0.02} &
  57.48 {\scriptsize ± 0.02} &
  57.31 {\scriptsize ± 0.01} &
  56.99 {\scriptsize ± 0.02} &
  57.02 {\scriptsize ± 0.03} &
  56.90 {\scriptsize ± 0.06} \\ \hline
ImageNet-1k &
  45.84 {\scriptsize ± 0.00} &
  46.23 {\scriptsize ± 0.01} &
  46.32 {\scriptsize ± 0.02} &
  46.09 {\scriptsize ± 0.00} &
  46.06 {\scriptsize ± 0.01} &
  46.19 {\scriptsize ± 0.00} &
  46.33 {\scriptsize ± 0.01} &
  45.89 {\scriptsize ± 0.01} &
  45.90 {\scriptsize ± 0.01} &
  45.65 {\scriptsize ± 0.02} \\ \hline
\end{tabular}%
}
\end{table*}

\begin{table}[t]
    \caption{Details of the image classification datasets on the ELEVATER benchmark.}\vspace{-0.2cm}
    \label{tab:elevaterdatasets}
    \centering
    \resizebox{\columnwidth}{!}{%
    \begin{tabular}{lccc}
        \hline
        Dataset & \#Labels & Test Size & Metric \\
        \hline
        Caltech-101~\cite{fei2004learning} & 101 & 6,084 & Mean-per-class \\
        CIFAR-10~\cite{krizhevsky2009learning} & 10 & 10,000 & Accuracy \\
        CIFAR-100~\cite{krizhevsky2009learning} & 100 & 10,000 & Accuracy \\
        Country-211~\cite{radford2021learning} & 211 & 21,100 & Accuracy \\
        DTD~\cite{cimpoi2014describing} & 47 & 1,880 & Accuracy \\
        EuroSAT~\cite{helber2019eurosat} & 10 & 5,000 & Accuracy \\
        FER-2013~\cite{goodfellow2013challenges} & 7 & 3,589 & Accuracy \\
        FGVC-Aircraft~\cite{maji2013fine} & 100 & 3,333 & Mean-per-class \\
        Food-101~\cite{bossard2014food} & 101 & 25,250 & Accuracy \\
        GTSRB~\cite{stallkamp2011german} & 43 & 12,630 & Accuracy \\
        Hateful-Memes~\cite{kiela2020hateful} & 2 & 500 & ROC AUC \\
        KITTI-Distance~\cite{fritsch2013new} & 4 & 711 & Accuracy \\
        MNIST~\cite{deng2012mnist} & 10 & 10,000 & Accuracy \\
        Oxford Flowers-102~\cite{nilsback2008automated} & 102 & 6,149 & Mean-per-class \\
        Oxford-IIIT Pets~\cite{parkhi2012cats} & 37 & 3,669 & Mean-per-class \\
        PatchCamelyon~\cite{veeling2018rotation} & 2 & 32,768 & Accuracy \\
        Rendered-SST2~\cite{radford2021learning} & 2 & 1,821 & Accuracy \\
        RESISC-45~\cite{cheng2017remote} & 45 & 25,200 & Accuracy \\
        Stanford-Cars~\cite{krause20133d} & 196 & 8,041 & Accuracy \\
        Pascal VOC-2007~\cite{everingham2010pascal} & 20 & 4,952 & 11-point mAP \\
        \hline
        Total & 1,151 & 192,677 & - \\
        \hline
        \end{tabular}
        }
\end{table}

\begin{figure}[h]
    \includegraphics[width=0.5\textwidth]{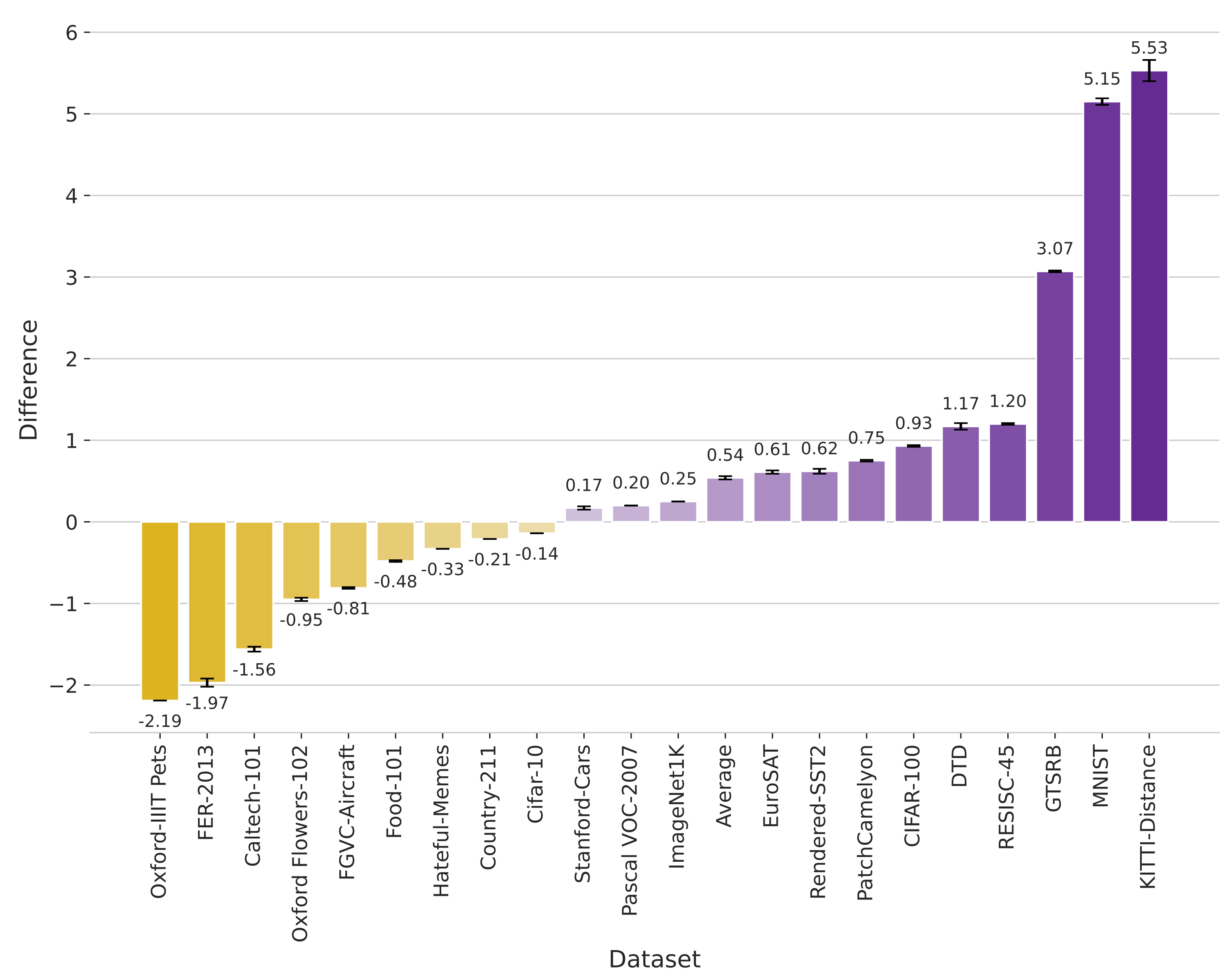}
    \caption{Difference between the \textsc{OpenCLIP} fine-tuned with CAPIVARA on CC3M, and the baseline (\textsc{OpenCLIP}), considering the ELEVATER benchmark and \text{ImageNet-1k}.}
    \label{fig:cc3m_10_gen}
\end{figure}

Figure~\ref{fig:cc3m_10_gen} presents the difference in performance between fine-tuning with CAPIVARA and the baseline, \textsc{OpenCLIP}. It can be noted that the majority of datasets exhibit positive differences in performance, indicating a favorable improvement over the baseline with CAPIVARA. Notably, the model trained with CAPIVARA led to substantial improvements of 5.53 and 5.15 \textit{pp.}~in two datasets, namely KITTI-Distance and MNIST, respectively. However, it is important to acknowledge instances where the performance of our model under this configuration falls short. Noteworthy cases include the Oxford-IIIT Pets dataset, encompassing 37 distinct breeds of cats and dogs, and the FER-2013 dataset, featuring a range of human emotional expressions. Also, our model presented a performance decline on these datasets, with respective decrements of 2.19 and 1.97 \textit{pp.} in comparison to the baseline.

\begin{figure}[p]
    \centering
    \includegraphics[width=0.435\textwidth]{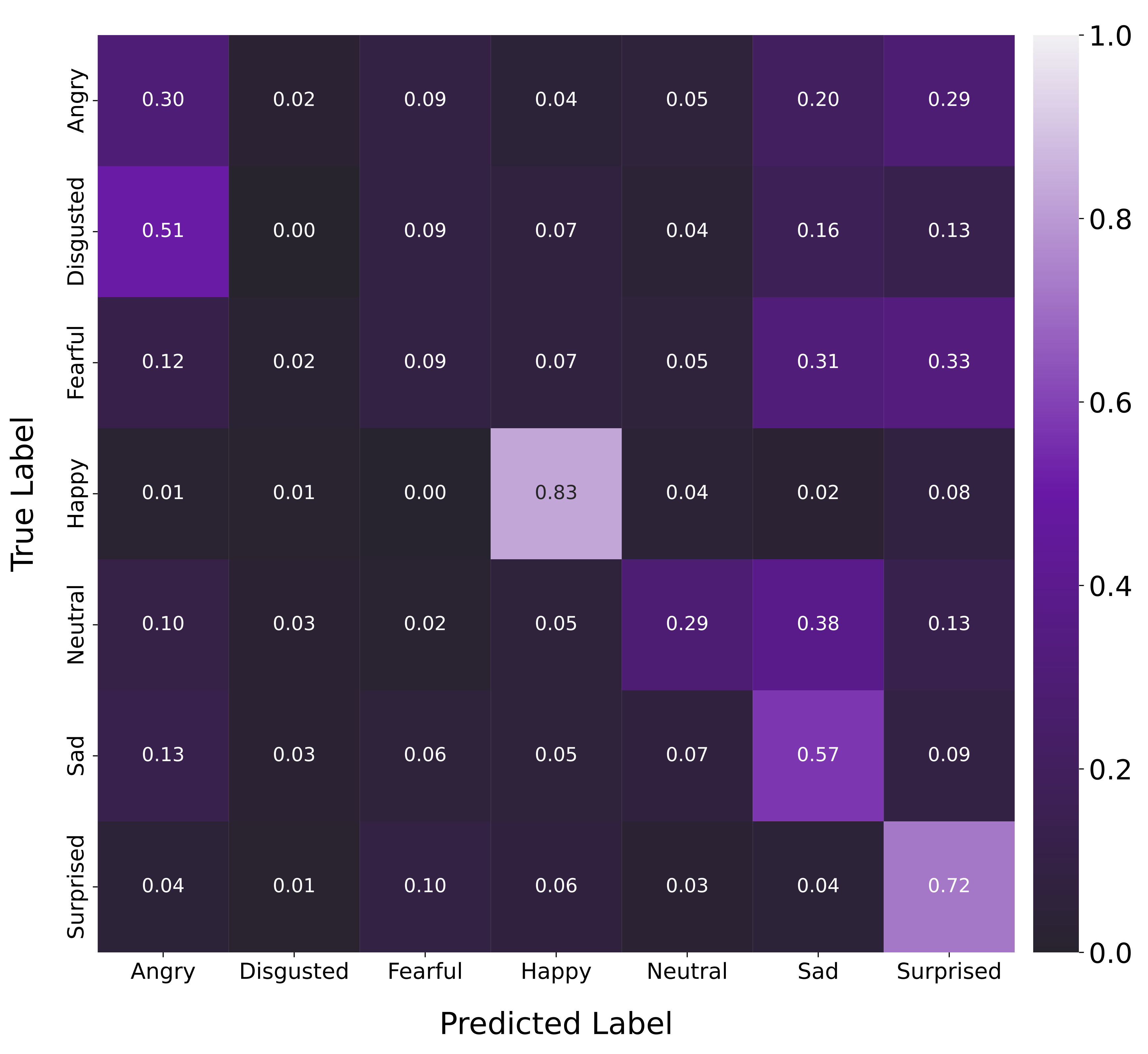}
    \caption{Normalized confusion matrix of the FER-2013 dataset for the \textsc{OpenCLIP} baseline model.}\vspace{-0.2cm}
    \label{fig:conf_matrix_fer_baseline}
\end{figure}

\begin{figure}[p]
    \centering
    \includegraphics[width=0.435\textwidth]{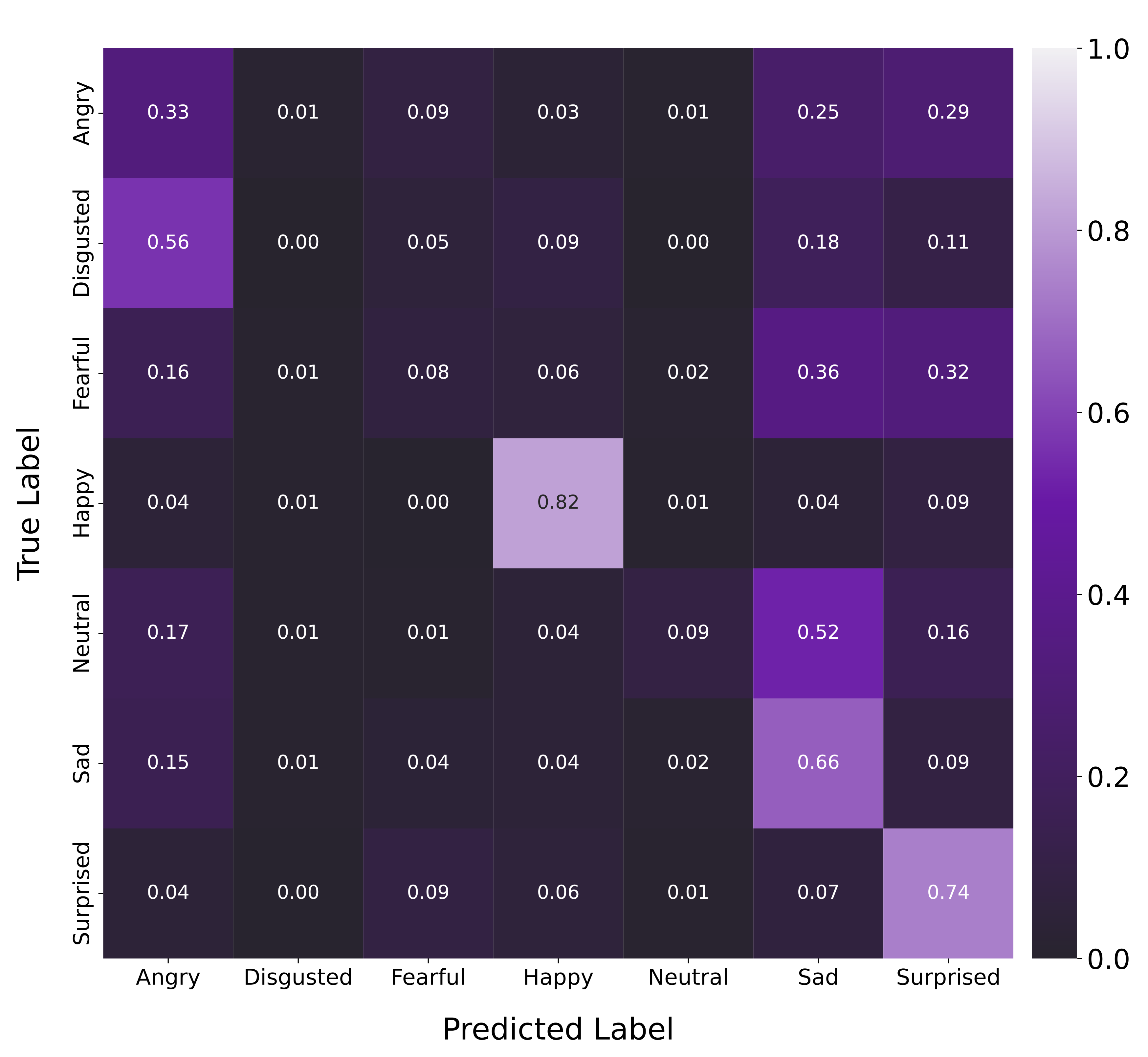}
    \caption{Normalized confusion matrix of the FER-2013 dataset for CAPIVARA.}\vspace{-0.2cm}
    \label{fig:conf_matrix_fer}
\end{figure}

\begin{figure*}[p]
\centering
    \includegraphics[width=0.75\textwidth]
    {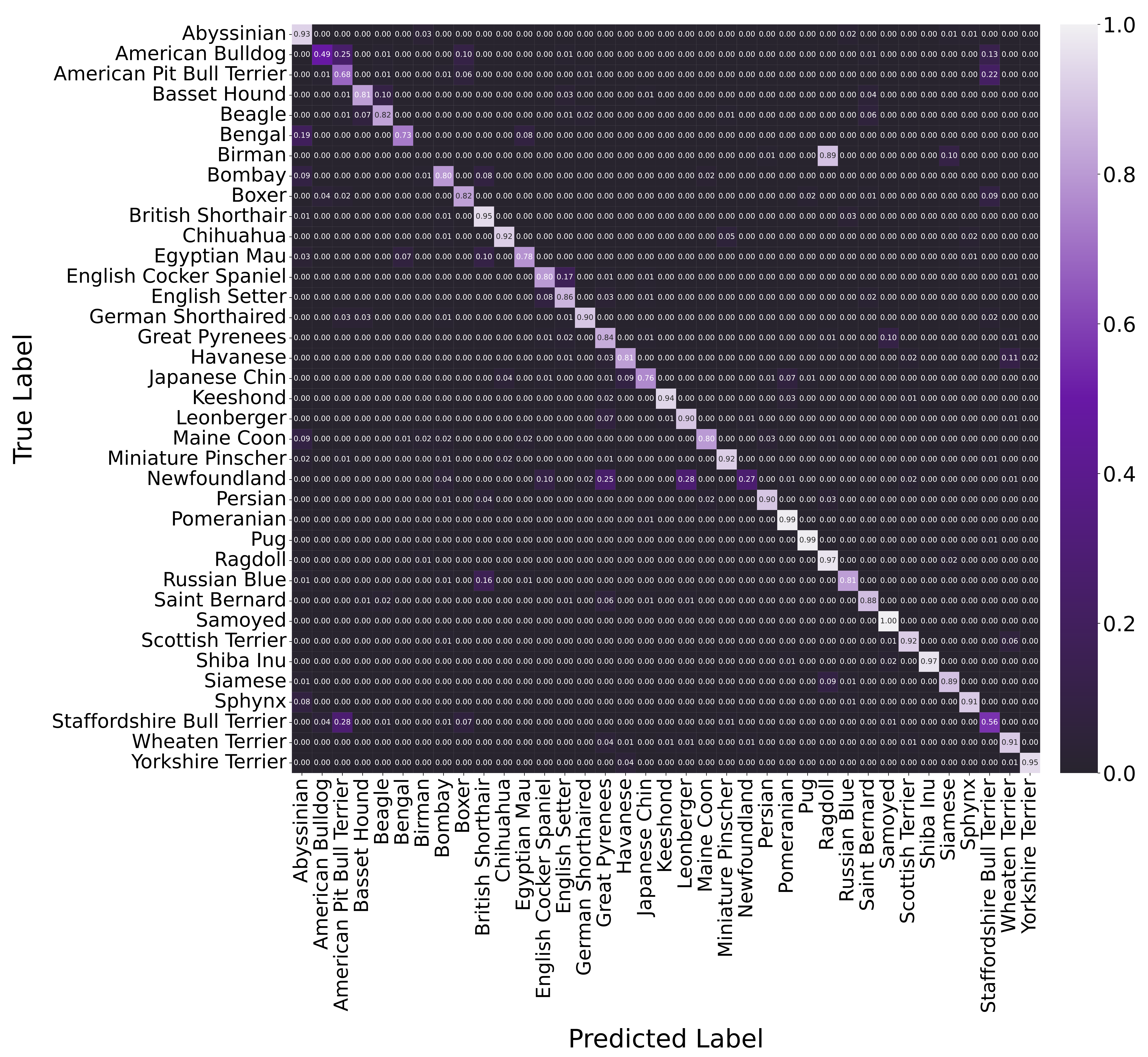}
    \caption{Normalized confusion matrix of the Oxford-IIIT Pets dataset for \textsc{OpenCLIP} baseline model.}
    \label{fig:conf_matrix_pets_baseline}
\end{figure*}

\begin{figure*}[p]
\centering
    \includegraphics[width=0.75\textwidth]
    {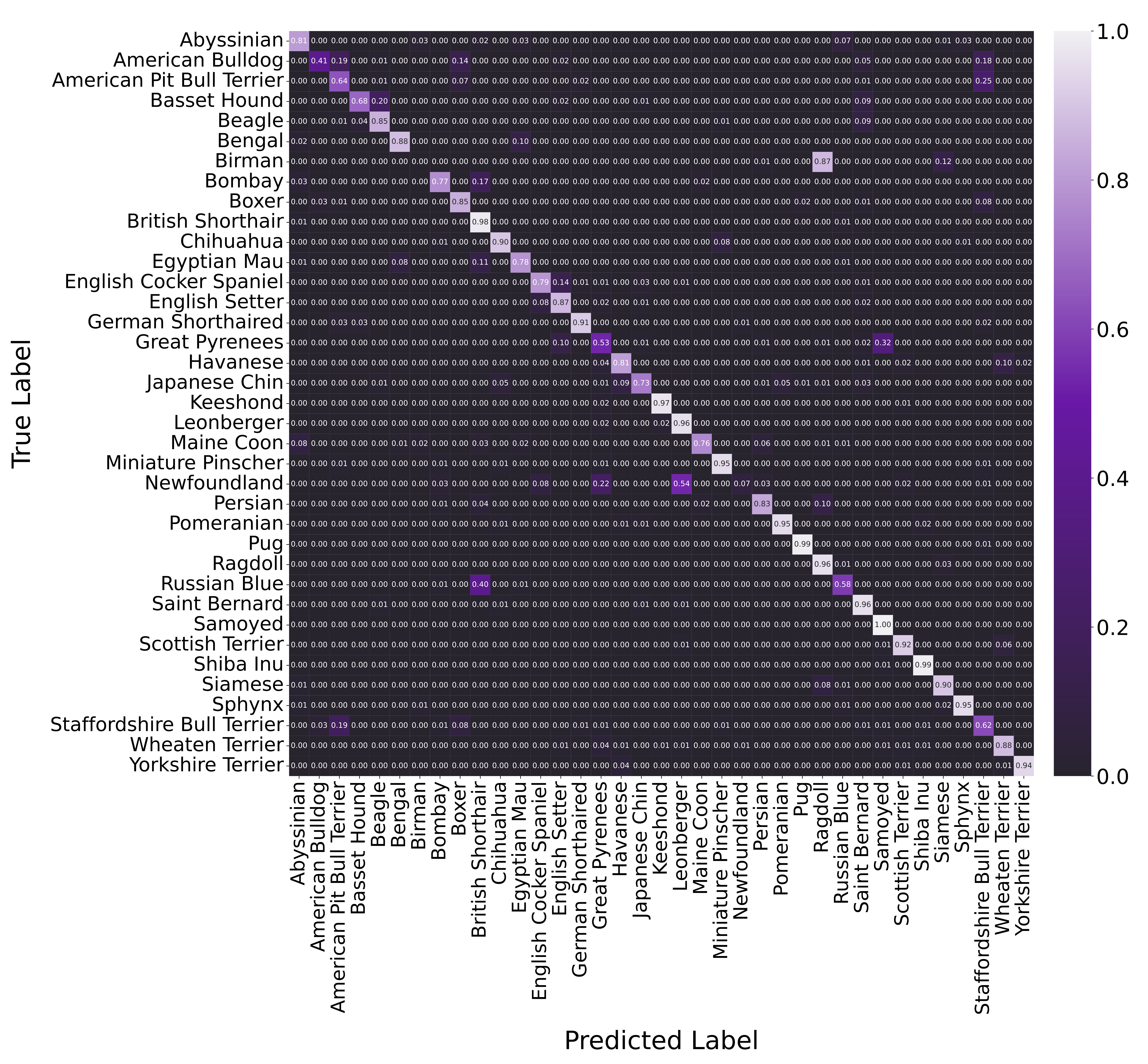}
    \caption{Normalized confusion matrix of the Oxford-IIIT Pets dataset for the \textsc{OpenCLIP} + Fine-tuning model with 10 generated annotations.}
    \label{fig:conf_matrix_pets}
\end{figure*}

Figures~\labelcref{fig:conf_matrix_fer_baseline,fig:conf_matrix_fer,fig:conf_matrix_pets_baseline,fig:conf_matrix_pets} offer a deeper dive into these observations, presenting normalized confusion matrices that provide granular insights into datasets where CAPIVARA underperformed the baseline. Specifically, Figures~\ref{fig:conf_matrix_fer_baseline} and~\ref{fig:conf_matrix_fer} unveil nuances in accurate and erroneous predictions within the Fer-2013 dataset. Notably, the baseline model excels in recognizing neutral expressions, while the fine-tuned model performs well in identifying expressions of sadness. However, the fine-tuned model is also more likely to confound emotions such as sadness and neutral expressions. Figures~\ref{fig:conf_matrix_pets_baseline} and~\ref{fig:conf_matrix_pets} present normalized confusion matrices for the Oxford-IIIT Pets Dataset, highlighting the fine-tuned model's tendency to amplify confusion between cat breeds British Shorthair and Russian Blue, as well as dog breeds Leonberger and Newfoundland, leading to reduced overall correctness. 

\subsection{Ablation Study}\label{sec:ablation_study}
\subsubsection{Impact of Multiple Captions \& Generated Caption Selection}
\label{sec:caption_selection}

To further validate the contributions of synthetic captions, we analyze the influence of multiple captions per image and how to select proper captions for each image. This latter aspect is related to BLIP2's hallucination, i.e., the model generates a text that does not match the associated image~\cite{xu2023lvlm}. The use of these synthetic annotations can introduce noise to the dataset. To address this issue, we implement the Captioning and Filtering (CapFilt)~\cite{blip_li_2022, blip2_li_2023} method with three different selection strategies: rank-based, threshold-based, and threshold-based $+$ near-duplication removal. All strategies rely on similarity scores produced by \textsc{OpenCLIP ViT-B/32 XLM-Roberta Base} model.

\begin{description}
\item \textbf{Rank-based:} We rank the synthetic captions along with the original descriptions based on the image-text similarity and select the top-k examples; in our tests, we adopted $k=5$.

\item \textbf{Threshold-based:} We select the texts among the original and generated captions based on their similarity to the associated image. Then, a caption is selected if the similarity between it and the image is greater than or equal to a given threshold; in this case, the threshold is~0.15.

\item \textbf{Threshold-based + near-duplication removal:} We first apply the threshold-based filter, and then we remove the near-duplicate captions using the algorithm described in Algorithm~\ref{alg:remove-duplication}, keeping a minimum of $k_{min}=3$ captions per image. Algorithm~\ref{alg:remove-duplication} first computes the text similarity matrix. Then, it computes the cost of removing a text $t_i$ as $c(t_i) =  \displaystyle\sum^B_{j=1} sim(t_i, t_j), \forall i \neq j$. At each step, it removes the text with the highest cost and updates the cost array. The algorithm stops when all similarity scores are lower than a given threshold or the minimum number of captions is reached. In this way, the algorithm can keep the maximum diversity among the texts. 
\end{description}

\newpage
\begin{lstlisting}[language=Python, label={alg:remove-duplication}, caption=Python-like pseudocode of near-duplicate text removal algorithm.]
# captions: image captions
# k_min: minimum number of texts to keep
# thr: maximum similarity between texts 
#      allowed

# Remove similar texts keeping the 
# maximum diversity among them
def remove_similar(captions, k_min=3, thr=0.3):
  if len(captions) < k_min:
    return captions

  sim_matrix = text_similarity(captions)
  n_texts = sim_matrix.shape[0]
  # set the cost in the diagonal to zero
  sim_matrix -= np.eye(n_texts)
  while not (sim_matrix <= thr).all() 
        and n_texts > k_min:
    # compute the cost to remove each 
    # text as sum of the similarity
    # between that text and all others.
    cost = sim_matrix.sum(axis=0)
  
    # remove the text with the highest 
    # cost 
    i = np.argmax(cost)
  
    # set the cost of the texts to be 
    # removed to zero
    sim_matrix[i, :] = 0
    sim_matrix[:, i] = 0
    n_texts -= 1

  # compute the final cost for all texts
  cost = sim_matrix.sum(axis=0)
  # all texts whose cost is zero will be 
  # removed
  remove_indices = np.where(cost==0)[0]
  # return the filtered texts
  return
   [caption
    for i,caption in enumerate(captions)
             if i not in remove_indices]
\end{lstlisting}

From a thorough analysis of the results exhibited in Table~\ref{table:caption_selection}, we note that none of the caption selection strategies significantly impacted the model performance. All strategies performed similarly to CAPIVARA with no caption selection. Specifically, the threshold-based caption selection strategy performed slightly better than the others but still in pair with CAPIVARA. This result suggests that BLIP2 is effective in generating captions related to images and, because of this, the caption selection methods did not affect the final performance. Nevertheless, Figure~\ref{fig:repeated_captions} and the results in Table~\ref{tab:multiple-captions} reveal that BLIP2 produces slightly different texts. Therefore, generating multiple captions per image has a limited effect on text augmentation. Note that adding 10 captions slightly improved compared to adding just one caption per image. Therefore, it is necessary to explore methods for generating more diverse texts, for instance, testing different sampling methods and other image captioning models, because we only used BLIP2 with default parameters.

\begin{table}[t]
\centering
\caption{Experimental results for caption selection strategies. In this table, ``threshold-based near-duplication'', ``threshold-based'', and ``rank-based'' refer to caption selection methods, whereas CAPIVARA does not consider any caption selection strategy. For each setting, we report the average and the standard deviation of mean recall.}\vspace{-0.2cm}
\label{table:caption_selection}
\resizebox{\columnwidth}{!}{%
\begin{tabular}{lcccccc}
\cline{2-7}
\multicolumn{1}{c}{}              & \multicolumn{2}{c}{Flickr30k} & \multicolumn{2}{c}{MS COCO} & \multicolumn{2}{c}{PraCegoVer} \\ \hline
\multicolumn{1}{l}{Method} & txt2img        & img2txt      & txt2img      & img2txt  & txt2img     & img2txt     \\ \hline
  \multicolumn{1}{l}{\begin{tabular}[l]{@{}l@{}}\textsc{OpenCLIP}\vspace{-0.1cm}\\(Baseline)\end{tabular}}    & 
76.23 & 
87.93 & 
52.62 & 
66.55 & 
65.36 & 
69.43 \\ \hline
\multicolumn{1}{l}{
\begin{tabular}[l]{@{}l@{}}\textsc{OpenCLIP}\\ + Fine-tuning\end{tabular}}         &
  \begin{tabular}[c]{@{}l@{}}78.42\vspace{-0.2cm}\\{\scriptsize ± 0.02}\end{tabular} & 
  \begin{tabular}[c]{@{}l@{}}90.02\vspace{-0.2cm}\\{\scriptsize ± 0.05}\end{tabular} & 
  \begin{tabular}[c]{@{}l@{}}54.77\vspace{-0.2cm}\\{\scriptsize ± 0.01}\end{tabular} & 
  \begin{tabular}[c]{@{}l@{}}70.06\vspace{-0.2cm}\\{\scriptsize ± 0.01}\end{tabular} & 
  \begin{tabular}[c]{@{}l@{}}63.79\vspace{-0.2cm}\\{\scriptsize ± 0.01}\end{tabular} & 
  \begin{tabular}[c]{@{}l@{}}60.10\vspace{-0.2cm}\\{\scriptsize ± 0.00}\end{tabular} \\ \hline
\multicolumn{1}{l}{\begin{tabular}[c]{@{}l@{}}Threshold-based\\ near-duplication\end{tabular}} &
  \begin{tabular}[c]{@{}l@{}}79.59\vspace{-0.2cm}\\{\scriptsize ± 0.01}\end{tabular} &
  \begin{tabular}[c]{@{}l@{}}90.02\vspace{-0.2cm}\\{\scriptsize ± 0.02}\end{tabular} &
  \begin{tabular}[c]{@{}l@{}}56.37\vspace{-0.2cm}\\{\scriptsize ± 0.01}\end{tabular} &
  \begin{tabular}[c]{@{}l@{}}71.14\vspace{-0.2cm}\\{\scriptsize ± 0.01}\end{tabular} &
  \begin{tabular}[c]{@{}l@{}}66.72\vspace{-0.2cm}\\{\scriptsize ± 0.01}\end{tabular} &
  \begin{tabular}[c]{@{}l@{}}65.33\vspace{-0.2cm}\\{\scriptsize ± 0.01}\end{tabular} \\ \hline
\multicolumn{1}{l}{Threshold-based} &
  \begin{tabular}[c]{@{}l@{}}79.65\vspace{-0.2cm}\\{\scriptsize ± 0.03}\end{tabular} &
  \begin{tabular}[c]{@{}l@{}}89.72\vspace{-0.2cm}\\{\scriptsize ± 0.02}\end{tabular} &
  \begin{tabular}[c]{@{}l@{}}56.39\vspace{-0.2cm}\\{\scriptsize ± 0.02}\end{tabular} &
  \begin{tabular}[c]{@{}l@{}}71.11\vspace{-0.2cm}\\{\scriptsize ± 0.02}\end{tabular} &
  \begin{tabular}[c]{@{}l@{}}66.77 \vspace{-0.2cm}\\{\scriptsize ± 0.01}\end{tabular} &
  \begin{tabular}[c]{@{}l@{}}65.47\vspace{-0.2cm}\\{\scriptsize ± 0.01}\end{tabular} \\ \hline  	
\multicolumn{1}{l}{Rank-based} &
  \begin{tabular}[c]{@{}l@{}}79.60\vspace{-0.2cm}\\{\scriptsize ± 0.01}\end{tabular} &
  \begin{tabular}[c]{@{}l@{}}89.13\vspace{-0.2cm}\\{\scriptsize ± 0.04}\end{tabular} &
  \begin{tabular}[c]{@{}l@{}}56.32\vspace{-0.2cm}\\{\scriptsize ± 0.01}\end{tabular} &
  \begin{tabular}[c]{@{}l@{}}70.64\vspace{-0.2cm}\\{\scriptsize ± 0.02}\end{tabular} &
  \begin{tabular}[c]{@{}l@{}}66.85\vspace{-0.2cm}\\{\scriptsize ± 0.00}\end{tabular} &
  \begin{tabular}[c]{@{}l@{}}65.96\vspace{-0.2cm}\\{\scriptsize ± 0.01}\end{tabular} \\ \hline 
\multicolumn{1}{l}{CAPIVARA} &
  \begin{tabular}[c]{@{}l@{}}79.56\vspace{-0.2cm}\\{\scriptsize ± 0.01}\end{tabular} &
  \begin{tabular}[c]{@{}l@{}}89.95\vspace{-0.2cm}\\{\scriptsize ± 0.04}\end{tabular} &
  \begin{tabular}[c]{@{}l@{}}56.27\vspace{-0.2cm}\\{\scriptsize ± 0.01}\end{tabular} &
  \begin{tabular}[c]{@{}l@{}}71.24\vspace{-0.2cm}\\{\scriptsize ± 0.01}\end{tabular} &
  \begin{tabular}[c]{@{}l@{}}66.40\vspace{-0.2cm}\\{\scriptsize ± 0.01}\end{tabular} &
  \begin{tabular}[c]{@{}l@{}}64.75\vspace{-0.2cm}\\{\scriptsize ± 0.01}\end{tabular} \\ \hline 
\end{tabular}%
}
\end{table}

\begin{table}[t]
\centering
\caption{Impact of multiple captions. This table presents the results of models trained with different numbers of synthetic captions translated into Portuguese. We report the average and the standard deviation of mean recall for each setting across Flickr30k, MS COCO, and PraCegoVer datasets.}\vspace{-0.2cm}
\label{tab:multiple-captions}
\resizebox{\columnwidth}{!}{%
\begin{tabular}{lcccccc}
\cline{2-7}
   & \multicolumn{2}{c}{Flickr30k} & \multicolumn{2}{c}{MS COCO} & \multicolumn{2}{c}{PraCegoVer} \\ \hline
\multicolumn{1}{l}{Method}                                                                      & txt2img        & img2txt    & txt2img      & img2txt  & txt2img      & img2txt      \\ \hline
  \multicolumn{1}{l}{\begin{tabular}[l]{@{}l@{}}\textsc{OpenCLIP}\vspace{-0.1cm}\\(Baseline)\end{tabular}}                                                                             & 76.23          & 87.93      & 52.62        & 66.55    & 65.36        & \textbf{69.43}        \\ \hline
\multicolumn{1}{l}{\begin{tabular}[l]{@{}l@{}}\textsc{OpenCLIP}\\ + Fine-tuning\end{tabular}} & 
\begin{tabular}[c]{@{}l@{}}78.42\vspace{-0.2cm}\\{\scriptsize ± 0.02}\end{tabular} & 
\begin{tabular}[c]{@{}l@{}}90.02\vspace{-0.2cm}\\{\scriptsize ± 0.05}\end{tabular} & 
\begin{tabular}[c]{@{}l@{}}54.77\vspace{-0.2cm}\\{\scriptsize ± 0.01}\end{tabular} & 
\begin{tabular}[c]{@{}l@{}}70.06\vspace{-0.2cm}\\{\scriptsize ± 0.01}\end{tabular} & 
\begin{tabular}[c]{@{}l@{}}63.79\vspace{-0.2cm}\\{\scriptsize ± 0.01}\end{tabular} &	
\begin{tabular}[c]{@{}l@{}}60.10\vspace{-0.2cm}\\{\scriptsize ± 0.00}\end{tabular}\\ \hline
\multicolumn{1}{l}{\begin{tabular}[c]{@{}l@{}}CAPIVARA \\+  10 synth. captions\end{tabular}} &
\textbf{\begin{tabular}[c]{@{}l@{}}79.56\vspace{-0.2cm}\\{\scriptsize ± 0.01}\end{tabular}} &
\textbf{\begin{tabular}[c]{@{}l@{}}89.95\vspace{-0.2cm}\\{\scriptsize ± 0.04}\end{tabular}} &
\textbf{\begin{tabular}[c]{@{}l@{}}56.27\vspace{-0.2cm}\\{\scriptsize ± 0.01}\end{tabular}} &
\textbf{\begin{tabular}[c]{@{}l@{}}71.24\vspace{-0.2cm}\\{\scriptsize ± 0.01}\end{tabular}} &
\textbf{\begin{tabular}[c]{@{}l@{}}66.40\vspace{-0.2cm}\\{\scriptsize ± 0.01}\end{tabular}} &
\begin{tabular}[c]{@{}l@{}}64.75\vspace{-0.2cm}\\{\scriptsize ± 0.01}\end{tabular} \\ \hline
\multicolumn{1}{l}{\begin{tabular}[c]{@{}l@{}}CAPIVARA \\+ 5 synth. captions\end{tabular}} & 
\begin{tabular}[c]{@{}l@{}}79.17\vspace{-0.2cm}\\{\scriptsize ± 0.02}\end{tabular} & 
\begin{tabular}[c]{@{}l@{}}90.72\vspace{-0.2cm}\\{\scriptsize ± 0.02}\end{tabular} & 
\begin{tabular}[c]{@{}l@{}}55.62\vspace{-0.2cm}\\{\scriptsize ± 0.01}\end{tabular} & 
\begin{tabular}[c]{@{}l@{}}70.95\vspace{-0.2cm}\\{\scriptsize ± 0.00}\end{tabular} & 
\begin{tabular}[c]{@{}l@{}}65.18\vspace{-0.2cm}\\{\scriptsize ± 0.01}\end{tabular} &	
\begin{tabular}[c]{@{}l@{}}62.14\vspace{-0.2cm}\\{\scriptsize ± 0.01}\end{tabular} \\ \hline
\multicolumn{1}{l}{\begin{tabular}[c]{@{}l@{}}CAPIVARA \\+ 1 synth. caption\end{tabular}}  & 
\begin{tabular}[c]{@{}l@{}}79.46\vspace{-0.2cm}\\{\scriptsize ± 0.01}\end{tabular} & 
\begin{tabular}[c]{@{}l@{}}90.02\vspace{-0.2cm}\\{\scriptsize ± 0.05}\end{tabular} & 
\begin{tabular}[c]{@{}l@{}}56.26\vspace{-0.2cm}\\{\scriptsize ± 0.01}\end{tabular} & 
\begin{tabular}[c]{@{}l@{}}71.27\vspace{-0.2cm}\\{\scriptsize ± 0.01}\end{tabular} & 
\begin{tabular}[c]{@{}l@{}}66.09\vspace{-0.2cm}\\{\scriptsize ± 0.01}\end{tabular} &	
\begin{tabular}[c]{@{}l@{}}63.95\vspace{-0.2cm}\\{\scriptsize ± 0.01}\end{tabular}\\ \hline
\end{tabular}%
}
\end{table}

\subsubsection{Impact of Increasing the Batch Size}

Among the different hyperparameters used to train the model, batch size has significant potential to improve model results. As batch size increases, more examples are observed per training step, and more examples might be discriminated by contrastive learning. Therefore, to determine the optimal batch size to use in our method, we conducted experiments fixing the number of steps in 5863 and varying this value considering our GPU memory limitation. We experimented three different batch sizes: 1000, 2816, and 4300. Each setting was tested with traditional fine-tuning and with CAPIVARA, the results are presented in Table~\ref{tab:batch-size}. 

Overall, we do not observe a significant gain in increasing the batch size. Intriguingly, in the context of CAPIVARA, the performance slightly improves across the datasets as we increase the batch size from 1000 to 2816. However, it declines when we use a batch size of 4300. For this reason, the CAPIVARA models were trained with an average batch size of 2816, while the optimized CAPIVARA models were trained with a batch size of 1000. This study shows that using smaller batches to train the optimized models does not result in significant loss. At the same time, it saves memory and training time.

\begin{table}[]
\caption{Comparison between different batch sizes in fine-tuning and CAPIVARA settings.}\vspace{-0.2cm}
\label{tab:batch-size}
\centering
\resizebox{\columnwidth}{!}{%
\begin{tabular}{cccccccc}
\cline{3-8}
\multicolumn{1}{l}{} &
  \multicolumn{1}{c}{} &
  \multicolumn{2}{c}{Flickr30k} &
  \multicolumn{2}{c}{MS COCO} &
  \multicolumn{2}{c}{PraCegoVer} \\ \hline
Method &
  \multicolumn{1}{c}{Batch size} &
  txt2img &
  img2txt &
  txt2img &
  img2txt &
  txt2img &
  img2txt \\ \hline
\multirow{3}{*}{\begin{tabular}[c]{@{}c@{}}OpenCLIP\\ + Fine-tunning\end{tabular}} &
  1000 &
  \begin{tabular}[c]{@{}c@{}}78.68\vspace{-0.2cm}\\{\scriptsize \textcolor{black}{± 0.02}}\end{tabular} &
  \begin{tabular}[c]{@{}c@{}}90.02\vspace{-0.2cm}\\{\scriptsize \textcolor{black}{± 0.02}}\end{tabular} &
  \begin{tabular}[c]{@{}c@{}}54.45\vspace{-0.2cm}\\{\scriptsize \textcolor{black}{± 0.01}}\end{tabular} &
  \begin{tabular}[c]{@{}c@{}}69.06\vspace{-0.2cm}\\{\scriptsize \textcolor{black}{± 0.01}}\end{tabular} &
  \begin{tabular}[c]{@{}c@{}}66.38\vspace{-0.2cm}\\{\scriptsize \textcolor{black}{± 0.01}}\end{tabular} &
  \begin{tabular}[c]{@{}c@{}}66.49\vspace{-0.2cm}\\{\scriptsize \textcolor{black}{± 0.02}}\end{tabular} \\
 &
  2816 &
  \begin{tabular}[c]{@{}c@{}}78.71\vspace{-0.2cm}\\{\scriptsize \textcolor{black}{± 0.02}}\end{tabular} &
  \begin{tabular}[c]{@{}c@{}}89.85\vspace{-0.2cm}\\{\scriptsize \textcolor{black}{± 0.02}}\end{tabular} &
  \begin{tabular}[c]{@{}c@{}}54.57\vspace{-0.2cm}\\{\scriptsize \textcolor{black}{± 0.00}}\end{tabular} &
  \begin{tabular}[c]{@{}c@{}}69.17\vspace{-0.2cm}\\{\scriptsize \textcolor{black}{± 0.03}}\end{tabular} &
  \begin{tabular}[c]{@{}c@{}}66.44\vspace{-0.2cm}\\{\scriptsize \textcolor{black}{± 0.01}}\end{tabular} &
  \begin{tabular}[c]{@{}c@{}}66.57\vspace{-0.2cm}\\{\scriptsize \textcolor{black}{± 0.01}}\end{tabular} \\
 &
  4300 &
  \begin{tabular}[c]{@{}c@{}}78.70\vspace{-0.2cm}\\{\scriptsize \textcolor{black}{± 0.01}}\end{tabular} &
  \begin{tabular}[c]{@{}c@{}}89.86\vspace{-0.2cm}\\{\scriptsize \textcolor{black}{± 0.02}}\end{tabular} &
  \begin{tabular}[c]{@{}c@{}}54.62\vspace{-0.2cm}\\{\scriptsize \textcolor{black}{± 0.04}}\end{tabular} &
  \begin{tabular}[c]{@{}c@{}}69.22\vspace{-0.2cm}\\{\scriptsize \textcolor{black}{± 0.02}}\end{tabular} &
  \begin{tabular}[c]{@{}c@{}}66.42\vspace{-0.2cm}\\{\scriptsize \textcolor{black}{± 0.05}}\end{tabular} &
  \begin{tabular}[c]{@{}c@{}}66.76\vspace{-0.2cm}\\{\scriptsize \textcolor{black}{± 0.19}}\end{tabular} \\ \hline
\multirow{3}{*}{\begin{tabular}[c]{@{}c@{}}CAPIVARA\\ + Opt.\\ (5863 steps)\end{tabular}} &
  1000 &
  \begin{tabular}[c]{@{}c@{}}79.71\vspace{-0.2cm}\\{\scriptsize \textcolor{black}{± 0.03}}\end{tabular} &
  \begin{tabular}[c]{@{}c@{}}90.51\vspace{-0.2cm}\\{\scriptsize \textcolor{black}{± 0.05}}\end{tabular} &
  \begin{tabular}[c]{@{}c@{}}55.36\vspace{-0.2cm}\\{\scriptsize \textcolor{black}{± 0.03}}\end{tabular} &
  \begin{tabular}[c]{@{}c@{}}69.58\vspace{-0.2cm}\\{\scriptsize \textcolor{black}{± 0.03}}\end{tabular} &
  \begin{tabular}[c]{@{}c@{}}67.00\vspace{-0.2cm}\\{\scriptsize \textcolor{black}{± 0.03}}\end{tabular} &
  \begin{tabular}[c]{@{}c@{}}68.01\vspace{-0.2cm}\\{\scriptsize \textcolor{black}{± 0.01}}\end{tabular} \\
 &
  2816 &
  \begin{tabular}[c]{@{}c@{}}79.81\vspace{-0.2cm}\\{\scriptsize \textcolor{black}{± 0.03}}\end{tabular} &
  \begin{tabular}[c]{@{}c@{}}90.65\vspace{-0.2cm}\\{\scriptsize \textcolor{black}{± 0.02}}\end{tabular} &
  \begin{tabular}[c]{@{}c@{}}55.56\vspace{-0.2cm}\\{\scriptsize \textcolor{black}{± 0.01}}\end{tabular} &
  \begin{tabular}[c]{@{}c@{}}69.64\vspace{-0.2cm}\\{\scriptsize \textcolor{black}{± 0.04}}\end{tabular} &
  \begin{tabular}[c]{@{}c@{}}67.07\vspace{-0.2cm}\\{\scriptsize \textcolor{black}{± 0.02}}\end{tabular} &
  \begin{tabular}[c]{@{}c@{}}68.14\vspace{-0.2cm}\\{\scriptsize \textcolor{black}{± 0.01}}\end{tabular} \\
 &
  4300 &
  \begin{tabular}[c]{@{}c@{}}79.87\vspace{-0.2cm}\\{\scriptsize \textcolor{black}{± 0.01}}\end{tabular} &
  \begin{tabular}[c]{@{}c@{}}90.63\vspace{-0.2cm}\\{\scriptsize \textcolor{black}{± 0.04}}\end{tabular} &
  \begin{tabular}[c]{@{}c@{}}55.63\vspace{-0.2cm}\\{\scriptsize \textcolor{black}{± 0.01}}\end{tabular} &
  \begin{tabular}[c]{@{}c@{}}69.70\vspace{-0.2cm}\\{\scriptsize \textcolor{black}{± 0.04}}\end{tabular} &
  \begin{tabular}[c]{@{}c@{}}67.08\vspace{-0.2cm}\\{\scriptsize \textcolor{black}{± 0.01}}\end{tabular} &
  \begin{tabular}[c]{@{}c@{}}68.19\vspace{-0.2cm}\\{\scriptsize \textcolor{black}{± 0.01}}\end{tabular} \\ \hline
\end{tabular}%
}
\end{table}

\subsection{Qualitative Analysis}

\begin{figure}[t]
    \includegraphics[width=0.5\textwidth]{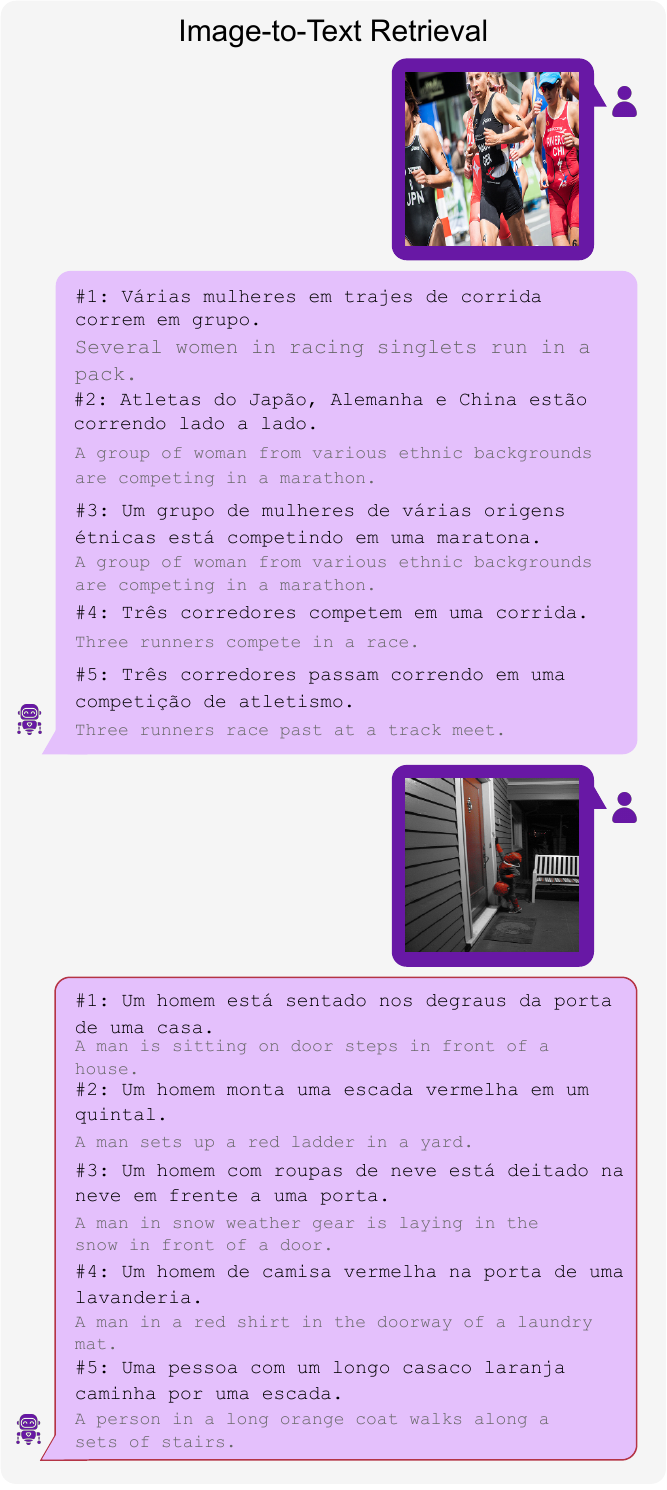}
    \caption{Examples of image-to-text retrieval using CAPIVARA + Opt.}
    \label{fig:I2T}
\end{figure}

\begin{figure}[t]
    \includegraphics[width=0.5\textwidth]{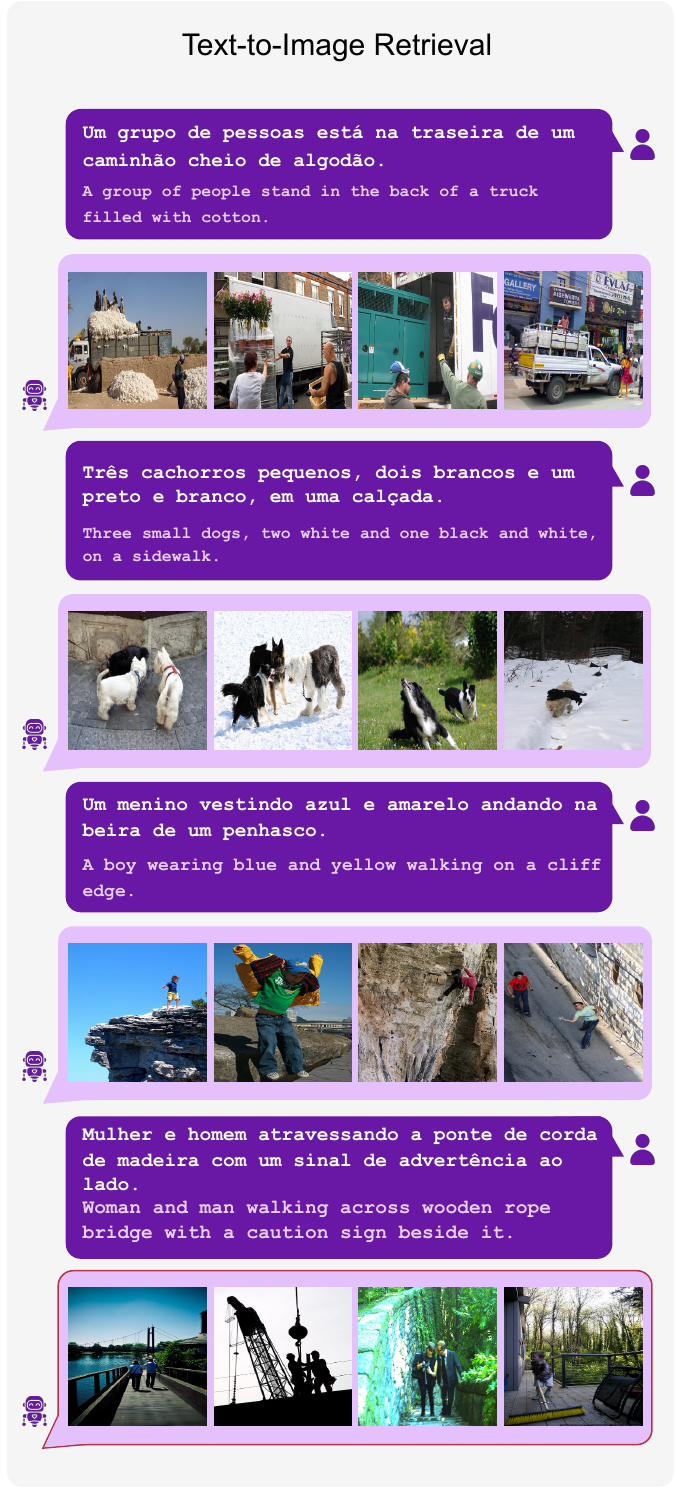}
    \caption{Examples of text-to-image retrieval using CAPIVARA + Opt.}
    \label{fig:T2I}
\end{figure}

We conducted experiments on Flickr30k for a qualitative analysis of the model's ability in cross-modal retrieval tasks, the outcomes are presented in Figures~\ref{fig:I2T} and~\ref{fig:T2I}. Figure~\ref{fig:I2T} shows the result of the image-to-text retrieval task, where the five texts in Portuguese more similar to a given image are retrieved by our model. For the first example, all the texts retrieved describe correctly the image content, which consists of a group of women running in a race. However, in the second example, none of the retrieved text matches the input image. It illustrates the limitations of our model. 

Similarly, we analyze qualitatively our model in text-to-image retrieval. In Figure~\ref{fig:T2I}, we present four examples of texts and the top-5 images more similar to each of them. We can see that overall the model ranks the correct images on the top.  Regarding the other images, although the scene representations match the texts, there is still a lack of details in the images that are not considered by the model, such as the number of people, objects, and colors. This can happen because there are no images that contain all elements from the text within the dataset, and it tries to retrieve the most similar images,  or by model limitations. Thus, in the last example, we present an instance in which the model fails. Given the text ``Woman and man walking across wooden rope bridge with a caution sign beside it.'', the model does not rank the expected image among the top-5 most similar. 

\subsection{Synthetic Captions Generated by BLIP2}

\begin{figure*}[t]
    \includegraphics[width=\textwidth]{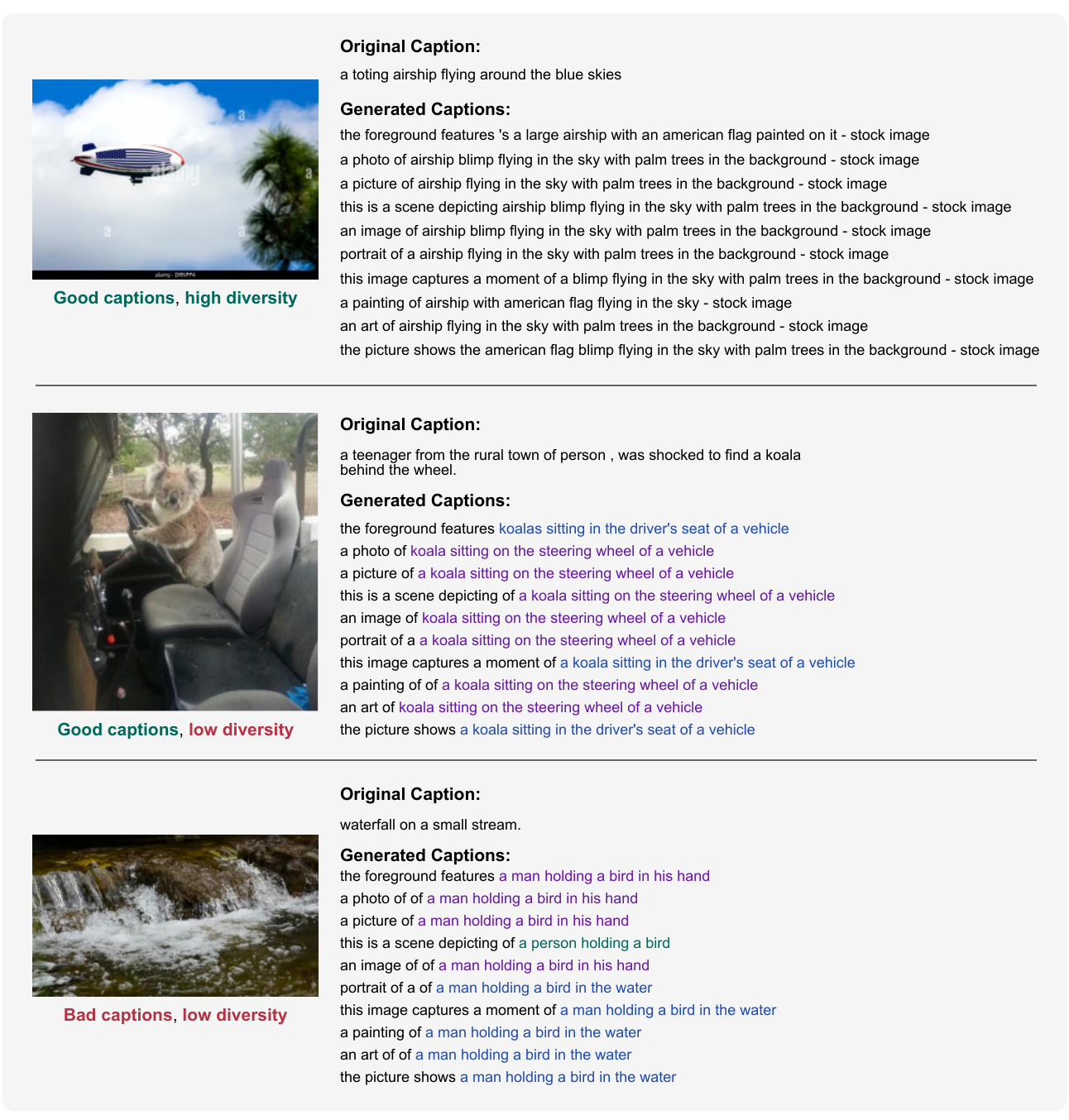}
    \caption{Examples of images with synthetic captions generated by BLIP2.}
    \label{fig:repeated_captions}
\end{figure*}

In the process of text augmentation, the BLIP2 model~\cite{blip2_li_2023} was used to generate new captions for the images. However, this model presents some issues regarding text generation. For example, it may generate text that does not match the image and repeat words. Several strategies have been used to mitigate these problems in our work. They are best described in Sec.~\ref{sec:method}. Figure~\ref{fig:repeated_captions} shows three images from CC3M along with their original caption and 10 captions generated with BLIP2.

The first image represents an example where the generated captions are good and diverse, as all captions correctly describe the image, there are no repeated words, and there is a high diversity of words used to describe the scene. The captions generally describe the image and add new elements to the description, although they still contain repetitive structures. In the second example, we present a scenario of good caption and low textual diversity. The captions describe the image, but there is a high level of repetition in the sentence structures. In the third example, we illustrate a case of badly generated captions and low textual diversity. In this example, the model not only shows a lot of word repetition, but also fails to represent the image, hallucinating.

\clearpage

\onecolumn
\subsection{Model Cards}
\label{sec:model_cards}
This section was done using the Model Cards for Model Reporting~\cite{mitchell2019model} tool.

\vspace{0.5cm}
\noindent\textbf{Model Details}
\begin{itemize}
    \item Developed by researchers from the Natural Language Processing Group of the Artificial Intelligence and Cognitive Architectures Hub -- H.IAAC.
    \item CAPIVARA, 2023, v1.
    \item CAPIVARA is a cost-efficient framework designed to enhance the performance of multilingual CLIP models in low-resource languages. 
    \item CAPIVARA augments text data using image captioning and machine translation to generate multiple synthetic captions in low-resource languages. The training pipeline is optimized with LiT, LoRA, and gradient checkpointing to alleviate the computational cost.
    \item More information can be found on CAPIVARA's official GitHub \url{https://github.com/hiaac-nlp/CAPIVARA}. 
    \item For further information or questions, please contact Sandra Avila \url{avilas@unicamp.br}.

\end{itemize}

\noindent\textbf{Intended Use}
\begin{itemize}
    \item Intended to be used for general tasks focused on finding a representation in a common space for texts and images. Examples of tasks are image-to-text and text-to-image retrieval and image classification. 
    \item Particularly intended for scientific researchers.
    \item Not intended to be used with aspects, positions, and cultural values from an under-represented region (e.g., Brazilian memes) due to the lack of representativeness of the datasets used for training. It cannot be used with long texts (more than 77 tokens).
\end{itemize}

\noindent\textbf{Factors}
\begin{itemize}
    \item Based on known problems with image and language models, potential relevant factors include groups for under-represented and minority people. In order to adapt the model to languages with low resources, texts were initially translated from English; thus, the model does not represent the cultural and geographical aspects of the countries that speak these target languages. The datasets used are made of texts collected from the Internet; therefore, the model may not perform as well for data collected from other sources and may carry biases from the original texts.
    
\end{itemize}

\noindent\textbf{Metrics}
\begin{itemize}
    \item Evaluation metrics include Mean Recall, representing the average recall value across the recall@K instances, where K = {1, 5, 10}, for cross-modal retrieval, which is the main task of CAPIVARA, and top-1 accuracy metrics for image classification task on ImageNet-1k. Moreover, the ELEVATER benchmark was used for the image classification task, and Appendix~\ref{sec:results_elevater_imagenet} provides the specific metrics used (see Table~\ref{tab:elevaterdatasets}). 

    \item  Each experiment was run three times, and the mean and standard deviation were reported for all experiments performed (see Section~\ref{sec:experiments}). 
  
\end{itemize}

\noindent\textbf{Quantitative Analyses}
\begin{itemize}
    \item Quantitative Analyses can be seen in Figure~\ref{fig:low-resource-performance} and Section~\ref{sec:experiments}.
\end{itemize}

\newpage
\noindent\textbf{Evaluation Data}
\begin{itemize}
    \item Evaluation data include Flickr30k, MS COCO, and PraCegoVer datasets for cross-modal retrieval task, and all 20 datasets from ELEVATOR benchmark and ImageNet-1k for image classification task (see Table~\ref{tab:full-results-classification}). 
    \item These datasets were chosen because they are the most widely used datasets in the literature, except for PraCegoVer. PraCegoVer is a dataset with images and texts originally in Portuguese that was used precisely to evaluate linguistic and cultural aspects present in the Portuguese language. (NOTE: Data originally in English that has been translated into the target language will be made available with the~model).
    \item See Section~\ref{sec:datasets} for more details about data preprocessing. 
\end{itemize}

\noindent\textbf{Training Data}
\begin{itemize}
    \item Training data was CC3M dataset.
    \item This dataset was chosen because of the amount of example data provided and the better quality of the data. In addition, our limited computing infrastructure for training the model was considered.
    \item See Section~\ref{sec:datasets} for more details about data preprocessing. 
    \item It is possible that the model was trained with data where group distributions are not homogeneous and, therefore, encoded some type of bias. 
\end{itemize}

\noindent\textbf{Ethical Considerations}
\begin{itemize}
    \item CAPIVARA does not deliberately use sensitive data in training. However, since it uses data collected from the Internet consisting of images and annotations about the image's content, it is possible that data with political, religious, or cultural positioning have been used.

    \item CAPIVARA does not generate any type of data that could pose a risk to human life. However, our model can be adapted for other specific tasks, e.g., image or text generation, which could contribute to generating false information and harming people.

    \item The model's training data was translated via Google Translate from English into the target language. This can lead to linguistic biases and a lack of representativeness for the target groups.
    
    \item CAPIVARA adopts training time optimizers, resulting in a smaller carbon footprint than traditional fine-tuning. Therefore, it presents a better financial and environmental alternative to improve the performance of pre-trained models.
\end{itemize}

\noindent\textbf{Caveats and Recommendations}
\begin{itemize}
    \item Further work is needed to assess the impact of adding more samples from the target language and how much this brings the performance of the target language closer to  English, which currently has the best performances. See Section~\ref{sec:conclusion} for more future works. 

    \item People and groups who do not have access to the Internet and, therefore, do not produce digital content were under-represented in the training set. However, CAPIVARA is intended to be applied to languages with low digital resources. CAPIVARA offers the technique to improve performance for low-resource languages, however, there is still a gap in performance between English texts and texts in low-resource languages. Future studies are required to improve performance for different languages and include cultural and linguistic aspects of the target language in the model.

    \item An ideal evaluation dataset would additionally include annotations made in the target language, which also considers cultural and linguistic aspects and has a background of minority and under-represented~groups.

    \item Current literature is constantly evaluating the ethical risks and impacts that vision and language models can have on society. Keeping up with this work is extremely important, as these studies can point to risks and negative impacts that have not yet been considered in this current version of Model Cards.

    \item Ideally, when using CAPIVARA as a base model for other applications, a study of the ethical impacts of the application should be carried out before it is implemented.

    \item It is highly recommended to read these Model Cards in conjunction with the article that introduces CAPIVARA, as the article contains detailed information on the entire life cycle of the proposed model.

\end{itemize}


\end{document}